\documentclass{article}

\usepackage{microtype}
\usepackage{graphicx}
\usepackage{subfigure}
\usepackage{booktabs} 

\usepackage{hyperref}

\usepackage[accepted,nohyperref]{icml2023}

\usepackage{amsmath}
\usepackage{amssymb}
\usepackage{mathtools}
\usepackage{amsthm}

\usepackage[capitalize,noabbrev]{cleveref}

\theoremstyle{plain}
\newtheorem{theorem}{Theorem}[section]

\theoremstyle{definition}
\newtheorem{definition}[theorem]{Definition}

\theoremstyle{remark}

\usepackage[textsize=tiny]{todonotes}

\usepackage[utf8]{inputenc} 
\usepackage[T1]{fontenc}    
\usepackage{hyperref}       
\usepackage{url}            
\usepackage{booktabs}       
\usepackage{amsfonts}       
\usepackage{nicefrac}       
\usepackage{microtype}      
\usepackage{xcolor}         
\usepackage[flushleft]{threeparttable}
\usepackage{bbm}
\usepackage{algorithm}
\usepackage{algorithmic}

\usepackage{graphicx}
\usepackage{sidecap}
\usepackage[sort&compress,numbers]{natbib}%
\usepackage{wrapfig}
\usepackage{bm}
\usepackage{relsize}

\newcommand{\squish}[1]{{#1\parfillskip=0pt\par}}

\usepackage{makecell}
\usepackage{mathtools}
\DeclareMathOperator{\argmax}{argmax}
\usepackage{multirow}

\icmltitlerunning{Adaptive Identification of Populations with Treatment Benefit in Clinical Trials}

\begin{document}

\twocolumn[
\icmltitle{Adaptive Identification of Populations with Treatment Benefit in Clinical Trials: Machine Learning Challenges and Solutions}



\icmlsetsymbol{equal}{*}

\begin{icmlauthorlist}
\icmlauthor{Alicia Curth}{yyy}
\icmlauthor{Alihan H\"uy\"uk}{yyy}
\icmlauthor{Mihaela van der Schaar}{yyy,comp}
\end{icmlauthorlist}

\icmlaffiliation{yyy}{Department of Applied Mathematics and Theoretical Physics, University of Cambridge, UK}
\icmlaffiliation{comp}{The Alan Turing Institute}

\icmlcorrespondingauthor{Alicia Curth}{amc253@cam.ac.uk}
\icmlkeywords{Machine Learning, ICML}

\vskip 0.3in
]



\printAffiliationsAndNotice{} 

\begin{abstract}
We study the problem of adaptively identifying patient subpopulations that benefit from a given treatment during a confirmatory clinical trial. This type of adaptive clinical trial has been thoroughly studied in biostatistics, but has been allowed only limited adaptivity so far. Here, we  aim to relax classical restrictions on such designs and investigate how to incorporate ideas from the recent machine learning literature on adaptive and online experimentation to make trials more flexible and efficient. We find that the unique characteristics of the subpopulation selection problem --  most importantly that (i) one is usually interested in finding subpopulations with \textit{any} treatment benefit (and not necessarily the single subgroup with largest effect) given a limited budget and that (ii) effectiveness only has to be demonstrated across the subpopulation \textit{on average} -- give rise to interesting challenges and new desiderata when designing algorithmic solutions. Building on these findings, we propose AdaGGI and AdaGCPI, two meta-algorithms for subpopulation construction. We empirically investigate their performance across a range of simulation scenarios and derive insights into their (dis)advantages across different settings.
\end{abstract}

\section{Introduction}
\squish{The existence of treatment effect heterogeneity across subgroups of patients poses a challenge to both the success of clinical trials testing the effectiveness of treatments \textit{and} the quality of treatment decisions in clinical practice when prescribing a drug that has been proven to be effective only for the average population \cite{thall2021adaptive, stallard2014adaptive, magnusson2013group}. Examples for such heterogeneity are ubiquituous in practice and include differences in treatment responses in cancer patients with specific mutations \cite{nahta2003her}, pyschiatric patients with different forms of depression \cite{fournier2010antidepressant} and stroke patients \cite{rosenblum2016group}. Motivated by this, the problem of discovering treatment effect heterogeneity using \textit{logged} experimental or observational data has received much attention in the recent machine learning (ML) literature \cite{bica2021real}, resulting in the adaptation of many supervised ML methods for post-hoc effect estimation  \cite{hill2011bayesian, wager2018estimation,  alaa2018limits, shalit2017estimating, curth2021inductive}. The \textit{active} counterpart to this problem, i.e. designing experiments (clinical trials) to actively discover subpopulations that respond well to a treatment, has received only limited attention in the ML literature thus far but is the focus of this paper.}

\squish{The biostatistics literature on adaptive clinical trials, on the other hand, has proposed and extensively studied the use of so-called \textit{adaptive enrichment designs}, which allow to change both enrolment criteria and the null hypothesis to be tested in a clinical trial based on interim data (see e.g. \cite{stallard2014adaptive, thall2021adaptive} and Appendix \ref{app:addtrial} for an overview). In such designs, the degree of adaptivity and flexibility is usually quite limited as the ability to adapt features is commonly restricted to a few pre-specified interim analysis points and the number of subgroups is  often very small (most often set to exactly two).}

\squish{In this paper, we consider a new approach to designing such adaptive enrichment trials and investigate whether and how it is possible to make them more flexible and efficient by adapting tools that were originally developed to solve pure exploration\footnote{Note that, as further discussed in Appendix \ref{app:bandits}, we focus on adapting ideas from the literature on \textit{purely explorative} bandits, which \textit{do not} trade off exploration with exploitation and are thus very different from prototypical explore-exploit bandit problems.} multi-armed bandits \cite{bubeck2009pure} and other adaptive experiments problems in the recent ML literature. We find that the problem of constructing subpopulations from subgroups in which a treatment has \textit{any positive} effect most closely resembles the \textit{good} arm identification (or thresholding bandit) problem studied in e.g. \cite{locatelli2016optimal, zhong2017asynchronous, tao2019thresholding, mukherjee2017thresholding, kano2019good, katz2020true} as there is no need to limit treatment prescription to the single subgroup with largest effect \cite{jennison2007adaptive}. Nonetheless, we argue that there are additional unique characteristics of our problem that may change how algorithmic solutions should be designed: (i) clinical trials operate under constraints on \textit{both} budget and confidence, (ii) budget is \textit{very limited} compared to e.g. online advertising settings, (iii) effectiveness only has to be demonstrated across a subpopulation \textit{on average} and (iv) required control of false discovery and power is stricter and more nuanced. Note that solutions for problems with some of these characteristics could be of independent interest in applications beyond the clinical trial context: e.g. (i) and (ii) may appear whenever one is looking to find \textit{any (single) good} candidate, solution or arm with high confidence as fast as possible, while (iii) appears when one only needs to identify \textit{a collection of arms} that works well on average. }

\begin{figure}[t]
    \includegraphics[width=0.99\columnwidth]{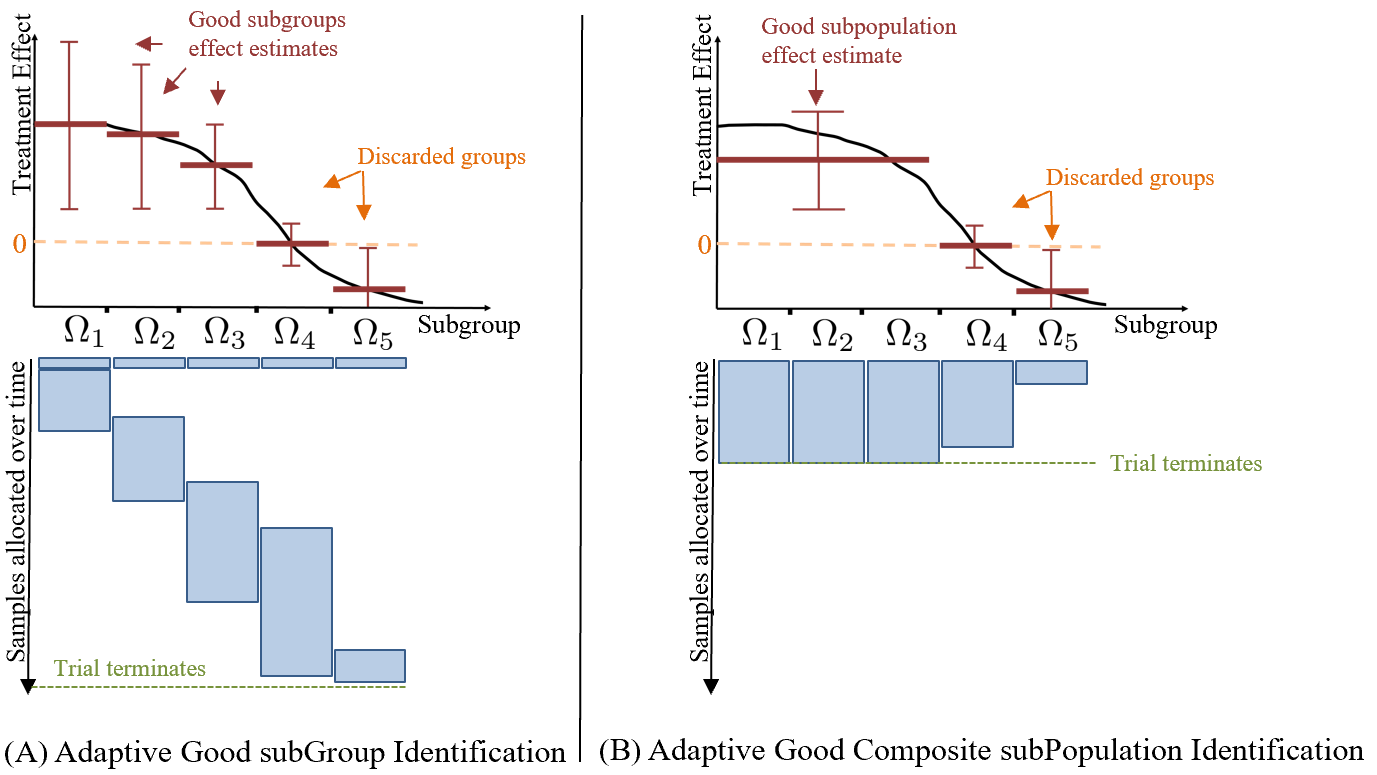}
    \vspace{-6pt}
    \caption{Overview of the two problem formulations and proposed solutions. (A) The adaptive good subgroup identification (AdaGGI) algorithm finds individual subgroups with treatment benefit through successive discovery. (B) The adaptive good composite subpopulation (AdaGCPI) algorithm finds a composite subpopulation by successively removing the subgroup with smallest effect until a positive average treatment effect is discovered.}\label{fig:overview}
    \vspace{-\baselineskip}
\end{figure}

\textbf{Contributions.}  We study the problem of adaptive identification of patient subpopulations that benefit from a treatment during a clinical trial through a ML lens. Note that our focus in this paper lies not primarily in developing novel ML methodology, but rather \textit{in formalizing and understanding our clinical trial problem and its inherent challenges as a novel ML problem}, then allowing us to explore how to best adapt existing solutions to our setting. In doing so, we hope to introduce relevant ML communities to a new application, through showcasing that this area is full of new ML problems, demanding constraints and interesting methodological challenges. We make three main contributions: \\ \squish{\textbf{(1) Problem formalization and understanding:} We focus on \textit{formalising},  \textit{contextualizing} and \textit{understanding} the population identification problem and its inherent challenges \textit{as a ML problem}. We discover two possible formulations of the problem which differ in terms of their characteristics and investigate how these give rise to different desiderata when designing algorithmic solutions. \mbox{\textbf{(2) Two new meta-algorithms:}} Building on these insights and ideas from the ML literature on adaptive experiments, we then propose a solution in form of a meta-algorithm for each scenario (see Fig. \ref{fig:overview}).
{\textbf{(3) Empirical Insight:}} We empirically investigate and provide insight into the (dis)advantages of either formulation and their solution through a range of simulation studies. Albeit not our primary objective, we believe that some of these empirical insights could be of independent interest to researchers studying the problem of good \textit{arm} identification in a small sample regime. }
\vspace{-6pt}

\section{Problem Setup}

Throughout, we adopt problem setting and notation similar to \cite{magnusson2013group}. Thus, we wish to run a clinical trial to establish efficacy of a novel drug (T) relative to an established control (C) in patient population $\Omega_0$. We assume further that $\Omega_0$ is made up of $K$ disjoint and prespecified subgroups $\Omega_1, \ldots, \Omega_K$ where $\textstyle \Omega_0 = \cup_{j\leq K}\Omega_j$, across which efficacy may be expected to differ, e.g. due to known biological pathways or evidence from earlier trials. Let $\theta_j$ denote the treatment's effect (relative to control) within subgroup $j$, and let $\pi_j$ denote the prevalence of subgroup $j$ in the population.

\textbf{Goal. }\squish{To ensure success of the clinical trial, we aim to adaptively construct a \textit{composite subpopulation} composed of a subset $\mathcal{S} \subseteq \mathcal{K}=\{1, \ldots, K\}$ of the full population with $\textstyle \Omega_{\mathcal{S}}= \cup_{i \in \mathcal{S}} \Omega_i$, in which the treatment is \textit{effective} on average (if any exists); we refer to such subpopulations as \textit{good}:}

\begin{definition}
    (Good Subpopulation) \upshape
    A subpopulation~$\mathcal{S}\subseteq\mathcal{K}$ is a \textit{good subpopulation} iff $\theta_{\mathcal{S}}=\sum_{i\in\mathcal{S}}\frac{\pi_i}{\sum_{j\in\mathcal{S}}\pi_j}\theta_i > 0$.
\end{definition}%
\vspace{-\baselineskip}\vspace{\parskip}

Generally, to maximise patient benefit, we would like to identify the \textit{largest} subpopulations in which the treatment is effective -- i.e. if $\theta_i\! > \!\theta_j\! > \!0$, we prefer $\mathcal{S}^{ij} = \{i, j\}$ over $\mathcal{S}^i=\{i\}$ even though $\textstyle\theta_{{S}^{ij}} < \theta_{{S}^{i}}$.

\textbf{Null hypotheses and problem types. } We consider a null scenario of no treatment effect, i.e. $\theta_0=0$, giving rise to two types of problems and associated null hypotheses. In Sec. \ref{sec:goodsubgroup}, we first identify individual good \textit{subgroups}, i.e. find subgroups for which we can reject the null hypothesis
\begin{equation}
  \textstyle  H_{0j}: \theta_j = 0
\end{equation}
for the one-sided alternative $ H_{aj}: \theta_j > 0$. Clearly, when composing a subpopulation by including only subgroups in which the individual null hypotheses have been rejected, i.e. $\mathcal{S}^a=\{j: \theta_j >0\}$, the subpopulation as a whole will have positive effect too, i.e. $\theta_{\mathcal{S}^a}>0$. We refer to this problem as the \textit{Good subGroup Identification} (GGI) problem.

\squish{Often, clinical trials are not powered to detect effects in subgroups seperately; instead (when more than two subgroups are considered), the focus is set on demonstrating \textit{average} effectiveness across a sub\textit{population} as in \cite{magnusson2013group}. We therefore consider a second setting in Sec. \ref{sec:goodpop}: here, we wish to identify a \textit{composite} subpopulation $\mathcal{S}$ for which we can prove that the treatment is effective \textit{on average}, i.e. reject}\vspace{-4mm}
\begin{equation}
\textstyle    H_{0\mathcal{S}}: \theta_\mathcal{S} = 0
\end{equation}
for the one-sided alternative $H_{a\mathcal{S}}:  \theta_\mathcal{S} > 0$. We refer to this problem as the \textit{Good Composite subPopulation Identification} (GCPI) problem. Note that the underlying requirement is strictly weaker than in the GGI problem as rejecting $H_{0\mathcal{S}}$ does not require rejecting $H_{0j}$ for every $j\in \mathcal{S}$. 

{\textbf{Familywise error rate control. } Regulatory agencies usually require the familywise error rate (FWER), i.e. the probability of a Type 1 error, to be controlled in clinical trials \cite{fda2019}. Formally, the FWER of an algorithm $\mathcal{A}$ for the set of problem instances $\mathcal{P}$ under consideration is defined as}
\begin{align*}
\text{\small FWER}(\mathcal{A}; \mathcal{P})= \sup\nolimits_{\rho \in \mathcal{P}} \mathbb{P}_\rho(\text{\small $\mathcal{A}$ rejects a true null hypothesis})
\end{align*}
and FWER-control at the level of $\alpha \in (0, 1)$ requires that $\textrm{FWER}(\mathcal{A}; \mathcal{P})\allowbreak \leq \alpha$. Further, we can write
\begin{align*}
    \text{\small FWER}_{\scriptstyle GGI}(\mathcal{A}; \mathcal{P}) &\leq \sup\nolimits_{\rho \in \mathcal{P}}\mathsmaller{\sum^K_{j=1}}  \mathbb{P}_\rho(\text{\small $\mathcal{A}$ rejects true } H_{0j}) \\
    \text{\small FWER}_{\scriptstyle GCPI}(\mathcal{A}; \mathcal{P}) &\leq \sup\nolimits_{\rho \in \mathcal{P}} \mathsmaller{\sum_{\mathcal{S} \subseteq \mathcal{K}}} \\[-3pt]
    &\hspace{21pt}\mathbb{P}_\rho(\text{\small $\mathcal{A}$ selects $\mathcal{S}$ and rejects true $H_{0\mathcal{S}}$})
\end{align*}

\vspace{-3pt}
{\textbf{Minimum relevant effect. } Clinical trials also aim to avoid Type 2 error (failure to detect a true positive effect). As the sample size needed to differentiate $\theta_0\!=\!0$ from $\theta_j\!>\!0$ scales as $\theta_j^{-2}$, trials often introduce a \textit{minimum clinically relevant difference} $\theta_{min}\!>\!\theta_0\!=\!0$ which a trial should be powered to detect \cite{copay2007understanding}. Thus, while not a hard requirement like FWER control, we aim to ensure that $\mathbb{P}(H_{0\mathcal{S'}} \textrm{ is not rejected } | \theta_{\mathcal{S}}\!=\!\theta_{min}) \!\leq\! \beta$ for at least some $\mathcal{S'}\! \subset \!\mathcal{S}$.}

{\textbf{Mode of environment interaction and data structure.} Throughout, we assume the stylized setting of an unlimited stream of patients available for recruitment from each subgroup, where outcomes are revealed to the algorithm immediately; we discuss possible extensions to more realistic scenarios in Appendix B. That is, at every time step $t \in \{1, \ldots,  B\}$, where $2B$ is the total patient budget of the trial, the algorithm selects a subgroup $J_t \in \mathcal{K}$ to enrol two patients from, which are then \textit{randomly} assigned to one of each treatment and control arm. This gives rise to control and treated outcome $Y^C_t, Y^T_t \in \mathcal{Y}$, which could be continuous ($\mathcal{Y}=\mathbb{R}$) or binary ($\mathcal{Y}=\{0, 1\}$), and  produces a dataset of tuples $\mathcal{D}_t=\{(J_{t'}, Y^C_{t'}, Y^T_{t'}\}_{t' \leq  t}$. We denote by $N_i(t)=\sum_{t' \leq t}\mathbbm{1}\{J_{t'} = i\}$ and $N_\mathcal{S}(t) = \sum_{t' \leq t}\mathbbm{1}\{J_{t'} \in \mathcal{S}\}$ the number of patient pairs enrolled from a subgroup or a subpopulation by time t, respectively.}

\textbf{Estimators \& Inference.} Given randomization and assuming \textit{no interference} between patients, we have that $\theta_j = \mathbb{E}[Y^T_t - Y^C_t| J_t = j]$, so that we can estimate treatment effects simply as 
\begin{equation}
\textstyle    \hat{\theta}_{j, N_j(t)}=\frac{\sum^t_{t'=1} \mathbbm{1}\{J_{t'} = j\}(Y^T_{t'} - Y^C_{t'})}{N_j(t)}
\end{equation}
Whenever all subgroups $i$ in a subpopulation $\mathcal{S}$ were drawn according to their relative prevalence $\textstyle \frac{\pi_i}{\sum_{j \in \mathcal{S}} \pi_j}$, we can also estimate $\textstyle \hat{\theta}_{\mathcal{S}, N_\mathcal{S}(t)}\! =\!\frac{\sum^t_{t'=1} \mathbbm{1}\{S_{t'} \in \mathcal{S}\}(Y^T_{t'} - Y^C_{t'})}{N_\mathcal{S}(t)}$. Note that the $\hat{\theta}_{j, N_j(t)}$ will generally not be unbiased for $\theta_j$ as the $J_t$ were selected in a data-adaptive manner (see e.g. \cite{nie2018adaptively, shin2019sample}).

{Finally, standard approaches to statistical inference will generally not be valid when experiments are stopped adaptively, and we need to account for possible bias due to continuous monitoring of experiments. To retain the ability to perform valid inference, we therefore also assume that we have access to some always-valid confidence intervals \cite{johari2015always}; that is, similar to \cite{jamieson2018bandit} we rely on existence of some function $\phi(t, \delta)$ which satisfies for any $\delta \!\in  \! (0, 1)$ that $\mathbb{P}(\cap^\infty_{t=1} \{|\hat{\theta}_{\mathcal{S}, t} - \theta_\mathcal{S}|  \leq \phi(t, \delta)\}) \! \geq  \! 1 \!- \! \delta$.  Our proposed meta-algorithms allow the use of any user-specified function  $\phi(t, \delta)$. As discussed further in Appendix \ref{app:details}, we follow \cite{jamieson2018bandit} in our experiments and instantiate it using Thm. 8 of \cite{kaufmann2016complexity} which shows that for mean-zero $\sigma^2_p$-(sub)gaussian variables $X_s$, $\mathbb{P}(\exists t  \! \in \! \mathbb{N} \!: \! \frac{\sum^t_{s=1}X_s}{t}> \sqrt{\frac{2\sigma_p^2\zeta(t, \delta)}{t}})  \!\leq \! \delta$ for $\zeta(t, \delta) \!= \!\log(1/\delta) \! +  \!3 \log \log(1/\delta) \!+ \!(3/2)\log \log(et/2)$ and $\delta\!\leq \! 0.1$. We can use $ \sqrt{\frac{2\sigma_p^2\zeta(t, \delta)}{t}}$ as $\phi(\cdot, \cdot)$ in our experiments due to the fact that (i) the difference between two $\sigma^2$-(sub)gaussian variables is $2\sigma^2$-(sub)gaussian and (ii) Bernoulli variables are $\frac{1}{4}$-subgaussian. Note that generally subgaussianity is satisfied by e.g. \textit{bounded} (and centered) outcomes $Y$.  Given that many medical outcomes and lab tests have bounded credible ranges, we therefore consider subgaussianity of outcomes to be a very reasonable assumption in this context.}

\section{Good Subgroup Identification}\label{sec:goodsubgroup}

\squish{We begin by studying the good subgroup identification (GGI) problem as it appears more closely related to problems studied in the recent ML literature. Recall that the GGI problem focusses on finding members of the set $\mathcal{H}_a = \{j: \theta_j > 0\}$, subject to FWER-$\alpha$-control and budget $2B$.}

\squish{\textbf{Related work.} If $\theta_j$ was the \textit{mean of a bandit arm} (instead of a subgroup treatment effect), GGI resembles problems that have been studied in the pure exploration literature as \textit{thresholding} bandit \cite{locatelli2016optimal, zhong2017asynchronous, tao2019thresholding, mukherjee2017thresholding}, \textit{good arm} identification (GAI) \cite{kano2019good, katz2020true} and hypothesis testing using bandits \cite{jamieson2018bandit, xu2021unified}.\footnote{{More typical exploration problems, e.g. best arm identification (e.g. \cite{audibert2010best, gabillon2012best, jamieson2014best}) are less relevant as our interest lies no in finding the group with \textit{the best} response to a drug \cite{jennison2007adaptive}; see also Appendix A.}} In addition to the difference in target of interest, a major difference between existing formulations and our problem are the constraints placed on an ideal solution. Unlike our problem, classical pure exploration problems usually operate \textit{either} under a fixed budget \textit{or} a fixed confidence constraint: For example, in \cite{locatelli2016optimal}'s thresholding bandit, which aims to classify \textit{all} arms as above or below a threshold, the fixed confidence setting requires \textit{all} classifications (both above and below the threshold) to be correct with fixed confidence $\delta$, while the fixed budget setting aims for the highest confidence in all classifications given a certain budget. All of \cite{kano2019good, katz2020true, jamieson2018bandit, xu2021unified} study a similar fixed confidence setting. Finally, \cite{atan2019sequential} is the only ML work we are aware of that studies good subgroup discovery in a clinical trial context -- they propose a Bayesian MDP-based design optimizing patient recruitment given a fixed budget but do not control Type I error rate of discoveries, which conceptually resembles a fixed-budget-only GAI setup. }

\subsection{Problem Characteristics and Design Considerations in the GGI Problem}

\squish{\textbf{Unique characteristics of the GGI objective. } Discovery in a clinical trial is usually subject to \textit{both} a budget \textit{and} FWER constraint (i.e. a fixed confidence constraint on \textit{each} discovery). Thus, instead of identifying \textit{all} good arms either under a fixed budget while maximising confidence as in \cite{locatelli2016optimal} or with fixed confidence while minimizing budget as in \cite{kano2019good, jamieson2018bandit}, we aim to maximise the number of arms that can be discovered with fixed confidence given a budget -- which is a combination of the fixed confidence and fixed budget setting that are usually considered separately. Additionally, the available budget is usually \textit{very limited} in clinical trials relative to e.g. online advertising applications commonly considered in the bandit literature. Due to both ethical and financial considerations, clinical trials usually operate in small sample regimes -- confirmatory phase 3 sample sizes usually lie between 300-3000 patients \cite{fdasamplesize}, which is orders of magnitude smaller than sample sizes considered in the ML literature. Finally, the distinction (or asymetry) between both confidence $\alpha$ and power $1-\beta$, and null threshold $\theta_0$ and minimum relevant effect $\theta_{min}$ is usually not found in e.g. GAI problems.}

\squish{\textbf{Design considerations.} The unique characteristics of the GGI objective give rise to a number of desiderata while designing algorithms: First, there is a \textit{need to focus on promising groups}, as budget is limited and to meet our objective it is \textit{not} necessary to make a judgement about \textit{all} subgroups immediately. Thus we should focus our attention on subgroups that look promising and leave subgroups with effects that are hard to distinguish from the null for last (this is the opposite strategy to thresholding bandit solutions  \cite{locatelli2016optimal, zhong2017asynchronous, tao2019thresholding} that focus explicitly on the arms that are hardest to identify\footnote{{If identifying \textit{all} good groups is desired, it matters how fast the last (most difficult) group is found; while our goal to identify \textit{many good groups quickly} necessitates early focus on `easier' ones.}}). Second, we may wish to \textit{limit the degree of exploration} and explore only so long until a promising good subgroup has been identified (this is unlike a \textit{best} arm identification problem where \textit{relative} quality of an arm matters which needs substantially more exploration to identify). Third, we may want to
\textit{focus on null hypotheses closest to rejection}, recognizing that for a successful trial, rejecting one null hypothesis at level $\alpha$ is better than having two hypotheses only close to rejection upon termination.}

\squish{The backbones of the fixed confidence algorithms for identifying good bandit arms with mean above a threshold proposed in \cite{jamieson2018bandit, kano2019good, katz2020true, xu2021unified} \textit{do} lend themselves to be adapted to our combined fixed confidence - fixed budget setting: these algorithms sequentially move arms from the \textit{active set} under exploration to a passive (output) set containing all good arms identified with fixed confidence thus far, and could in principle solve our fixed budget setting by simply stopping testing additional arms once the budget is reached. Below, we discuss this approach and our modifications in more detail.}

\subsection{AdaGGI: A Meta-algorithm for Good Subgroup Identification}\label{sec:adaggi}

{We propose AdaGGI, an \textit{Ada}ptive \textit{G}ood sub\textit{G}roup \textit{I}dentification meta-algorithm, presented in Alg. \ref{alg:ggi}. As described in detail below, each iteration consists of (i) choosing a subgroup $J_t$ to enrol using an exploration (sampling) strategy $\mathcal{E}$, (ii) subsequently screening for new good subgroups using an $\alpha$-dependent identification criterion $\mathcal{I}$ and (iii) removal of any groups demonstrating no minimum benefit using a $(\beta, \theta_{min})$-dependent removal criterion $\mathcal{R}$.} 

\begin{figure}[t]
\vspace{-6pt}
\begin{algorithm}[H]
    \centering
    \caption{AdaGGI}\label{alg:ggi}
    \footnotesize
    \begin{algorithmic}[1]
\REQUIRE {$\alpha, \beta \!\in\! (0, 1)$, $\theta_{min}\! > \!0$, budget $B$, initial samples $n_0$, \\ sampling rule $\mathcal{E}$, identification rule $\mathcal{I}$, removal rule $\mathcal{R}$}
\STATE Initialise: $\mathcal{A}_{Kn_0} = \mathcal{K}$; $\forall j\in \mathcal{K}$, sample $n_0$ times, \\ set $D_{Kn_0}=\{(S_{t'}, Y^C_{t'}, Y^T_{t'}\}_{t' \leq  Kn_0}$
\FOR{$t \in \{Kn_0 + 1, B\}$}
\STATE  Choose subgroup $J_t = \mathcal{E}(D_{t-1}, \mathcal{A}_{t-1})$ to enrol,\\set $\mathcal{D}_t = \mathcal{D}_{t-1} \cup (S_t, Y^C_t, Y^T_t)$
\STATE  Identify good subgroups $\mathcal{S}_t= \mathcal{S}_{t-1} \cup \mathcal{I}(\mathcal{D}_t, \alpha)$,
\\ set $\mathcal{A}_{t} = \mathcal{K} \setminus \mathcal{S}_t$
\STATE  Remove bad groups: $\mathcal{A}_{t} \!=\! \mathcal{K} \setminus  \mathcal{R}(\mathcal{D}_t, \theta_{min}, \beta)$
\STATE  If $\mathcal{A}_{t}= \emptyset$, \textbf{Output:} \texttt{True} if |$\mathcal{S}_B$|>0, $\mathcal{S}_B$
\ENDFOR
\STATE  \textbf{Output:} \texttt{True} if |$\mathcal{S}_B$|>0, $\mathcal{S}_B$
    \end{algorithmic}
\end{algorithm}
\vspace{-.5cm}
\end{figure}

\squish{\textbf{Sampling strategies $\mathcal{E}$: Finding good arms fast.} The established choice for sampling (exploration) strategy $\mathcal{E}$ in the GAI literature  \cite{kano2019good, jamieson2018bandit, katz2020true} appears to be to use an optimistic upper-confidence bound (UCB) approach, i.e. 
\begin{equation*}
    \mathcal{E}_{UCB}(D_{t\!-\!1}, \mathcal{A}_{t\!-\!1})\! =\! \arg \max_{j \in \mathcal{A}_{t\!-\!1}} \hat{\theta}_{j, N_j(t\!-\!1)}\! + \phi(N_j(t\!-\!1), \alpha)
\end{equation*} However, this strategy does not necessarily \textit{exploit} accumulated knowledge by repeatedly sampling a subgroup whose null is close to being rejected; in fact, as $\phi(t, \delta)$ shrinks with increasing $t$, we suspect that $\mathcal{E}_{UCB}$ may encourage frequent switching between subgroups when the effects in multiple good subgroups are similar which may lead to no null being rejected when budget is very limited.}

\squish{Therefore, we explore the use of two new sampling strategies for this problem. As we discuss below, identification using $\mathcal{I}(\cdot)$ will rely on the criterion $\mathbbm{1}\{\hat{\theta}_{j, N_j(t)} - \phi(N_j(t), \epsilon)>0\}$ for some $\epsilon \in (0,1)$; therefore, sampling according to the best lower confidence bound (LCB) would correspond to selecting arms that appear most promising for early identification, i.e. be more \textit{exploitative}. Thus, we also consider using}
\vspace{-.5cm}
\begin{equation*}
 \mathcal{E}_{LCB}(D_{t\!-\!1}, \mathcal{A}_{t\!-\!1}) = \arg \max_{j \in \mathcal{A}_{t-1}} \hat{\theta}_{j, N_j(t\!-\!1)} \!- \phi(N_j(t-1), \alpha)
\end{equation*} Because this strategy conversely may risk \textit{getting stuck} on a subgroup which only \textit{appeared} good early on, we consider a final strategy  $\mathcal{E}_{LUCB}(D_{t-1}, \mathcal{A}_{t-1})= \mathcal{E}_{LCB}(D_{t-1}, \mathcal{A}_{t-1}) \cup \mathcal{E}_{UCB}(D_{t-1}, \mathcal{A}_{t-1})$, allowing enrolment from two subgroups whenever sampling according to UCB and LCB disagree (thus $t$ increases by 2).

\squish{\textbf{Identification criterion: Ensuring FWER control.} Our identification criterion needs to ensure that $FWER_{GGI} \leq \alpha$ by adjusting for the fact that we perform \textit{multiple} hypothesis tests. As we consider only a moderate number of subgroups $K$, we rely on a simple Bonferroni correction here and use  
\begin{equation*}\mathcal{I}^K_{BF}(\mathcal{D}_t, \alpha) = \{j \in \mathcal{K}: \hat{\theta}_{j, N_j(t)} - \phi(N_j(t), \frac{\alpha}{K})>0\}
\end{equation*}
which controls FWER as 
\begin{equation*}
    \sum_{j \in \mathcal{K}: \theta_j = 0} \mathbb{P}(\cap^\infty_{t=1} \{\hat{\theta}_{j, t} - \theta_j  > \phi(t, \frac{\alpha}{K})\} \leq K \frac{\alpha}{K}
\end{equation*} To create tighter confidence bounds in settings  where \textit{many} null hypotheses are false and recycling $\alpha$ from previously rejected hypotheses is thus possible, one could implement more sophisticated strategies based on the adapted Benjamini-Hochberg procedure from \cite{jamieson2018bandit}, or other $\alpha$-investing approaches such as those discussed in \cite{tian2021online}.}

\textbf{Removal criterion: Focusing on significant effects. } We employ removal criterion 
\begin{equation*}
    \mathcal{R}_{fut}(\mathcal{D}_t,\allowbreak \theta_{min},\allowbreak \beta)\allowbreak\! =\!\allowbreak \{j\! \in\! \mathcal{K}\!:\! \hat{\theta}_{j, N_j(t)} + \phi(N_j(t), \beta)\!<\!\theta_{min}\}
\end{equation*} This ensures that subgroups can be removed early for \textit{futility} while power to detect a clinically relevant effect is preserved. Note that this ensures that the burden of proof to discard a bad subgroup can be much lower than what is needed to identify it as good. This differs from the recent GAI literature, where arms are either discarded and accepted using the same threshold/confidence \cite{kano2019good} or not discarded at all \cite{jamieson2018bandit, katz2020true}.

\section{Good Composite Subpopulation Identification}\label{sec:goodpop}

Instead of finding good subgroups \textit{separately} as before, we now move to the Good Composite subPopulation Identification (GCPI) problem which considers finding a good \textit{composite} subpopulation directly, i.e. finding $\mathcal{S}\subseteq \mathcal{K}$ such that $\textstyle \theta_\mathcal{S} = \sum_{i \in \mathcal{S}} \frac{\pi_i}{\sum_{j \in \mathcal{S}} \pi_j} \theta_i > 0$. Intuitively, this should be \textit{easier} to solve -- i.e. we would expect a smaller sample size to be required for a trial to be successful: given $\mathcal{S}$, rejection of $H_{0\mathcal{S}}$ is a strictly weaker requirement than rejecting all constituent elementary null hypotheses separately and it should be possible to share statistical strength (i.e. exploit larger sample size) \textit{across} subgroups contained in $\mathcal{S}$.

\textbf{Related work.} Most work from the adaptive enrichment clinical trial literature appears to solve a simplified version of the GCPI problem, where $\mathcal{K}=\{1, 2\}$ and initially patients from both subgroups are enrolled. At either a single (e.g. \cite{stallard2014adaptive, jenkins2011adaptive, friede2012conditional}) or multiple (e.g. \cite{rosenblum2016group, rosenblum2016multiple}) prespecified interim analysis points it is then possible to discontinue either subgroup, where decisions are usually based on precalculated (normal) stopping boundaries. The setting considered in \cite{magnusson2013group} is most similar to our setup as no restrictions are placed on $K$: here, the choice of subgroups to include in the selected subpopulation $\mathcal{S}$ is \textit{fixed} at the first interim analysis and all subsequent analyses allow only early termination of the \textit{entire} subpopulation based on efficacy/futility error-spending boundaries which are calculated based on the assumption that all $\theta_j \geq 0$ (i.e. negative effects are not allowed). From a bandit perspective, the GCPI problem can be interpreted as a generic \textit{combinatorial bandit} problem \cite{chen2014combinatorial, gabillon2016improved}, where each subpopulation could be seen as a \textit{super-arm}; however, to the best of our knowledge no existing solutions exploit the idea of sharing statistical strength across arms by pooling samples and solutions derived from e.g. \cite{chen2014combinatorial, gabillon2016improved} would therefore resemble our GGI solution.

\subsection{Unique Problem Characteristics and Design Considerations in the GCPI Problem}

\textbf{Unique characteristics of the GCPI objective.} Relative to GGI, we consider two additional features key to the GCPI problem: On the one hand, the weaker requirement of identification of a positive \textit{average} effect should make it possible to share statistical strength \textit{across} subgroups, which may make the problem \textit{easier}. On the other hand, while the GGI problem has only $K$ subgroups with associated hypotheses to consider, the subpopulation construction problem is \textit{combinatorial} and there are $2^K$ possible subpopulations and null hypotheses, possibly making the problem \textit{harder}.

\squish{\textbf{Design considerations. } While the need to identify single groups fast in the GGI problem led us to consider highly non-uniform sampling  schemes, the possibility to share statistical strength across subgroups in the GCPI problem makes \textit{successive elimination} algorithms \cite{even2002pac, audibert2010best}, which uniformly sample all subgroups that have not yet been eliminated for futility, a more attractive alternative: intuitively speaking, if all subgroups had exactly the same (positive) effect, uniformly allocating samples across all groups would lead to rejection of the \textit{full population} composite null hypothesis using the same expected number of samples that the GGI problem would need to identify a \textit{a single} group. Note that such potential efficiency of successive elimination in the GCPI problem stands in stark contrast to what has been observed for the \textit{best arm} identification problem, where UCB-style algorithms empirically dominate successive elimination algorithms which are too wasteful in that context (see e.g. \cite{jamieson2014lil}). Further, successive elimination has the inherent advantage that it substantially limits the number of subpopulations (and associated null hypotheses) the algorithm will consider: if subgroups are irreversibly eliminated one-by-one, an algorithm will consider at most $K$ (nested) subpopulations.}

\begin{figure}[t]
\begin{algorithm}[H]
    \centering
    \caption{AdaGCPI}\label{alg:gpi}
    \footnotesize
    \begin{algorithmic}[1]
\REQUIRE $\alpha, \beta \in (0, 1)$, $\theta_{min} > 0$, budget $B$, \\
identification rule~$\mathcal{I}$, removal rule $\mathcal{R}$
\STATE Initialise: $\mathcal{A}_{1} = \mathcal{K}$, set $\mathcal{D}_0=\emptyset$, $t=0$
\WHILE{$t < B$}
\STATE Sample each $j\!\in\! \mathcal{A}_t$, obtain $\mathcal{D}'\!=\! \{j, Y^C_{t+j}, Y^T_{t+j}\}_{j \in \mathcal{A}_t}$,
\\set $t\gets t+|\mathcal{A}_t|$, update $\mathcal{D}_{t}$ with $\mathcal{D}'$. 
\STATE Test for positive effect in current population $\mathcal{I}(\mathcal{D}_t, \alpha)$: 
\\ if detected, \textbf{Output:} \texttt{True}, $\mathcal{A}_t$
\STATE Remove bad groups: $\mathcal{A}_t = \mathcal{K} \setminus  \mathcal{R}(\mathcal{D}_t, \theta_{min}, \beta)$ and remove their samples from $\mathcal{D}_t$
\STATE If $\mathcal{A}_t= \emptyset$, \textbf{Output:} \texttt{False}, $\emptyset$
\ENDWHILE
\STATE \textbf{Output:} \texttt{False}, $\emptyset$
    \end{algorithmic}
\end{algorithm}
\vspace{-\baselineskip}
\end{figure}

\begin{figure*}[t]
    \centering
    \includegraphics[width=0.999\textwidth]{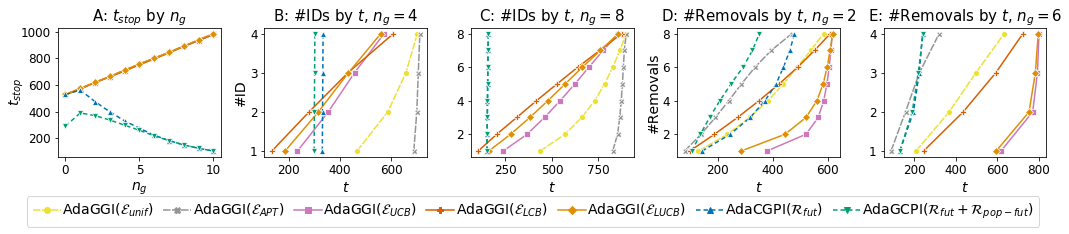}%
    \vspace{-10pt}%
    \caption{Results describing time until (A) termination, (B\&C) identification of good groups and (D\&E) removal of bad groups (1000 replications). (A): Time to termination $t_{stop}$ by \# of good groups $n_{g}$. (B\&C): \# of good group identifications over time, for $n_g\!=\!4$ (B) and $n_g\!=\!8$ (C). (D\&E): \# of removals of bad groups over time, for $n_g\!=\!2$ (D) and $n_g\!=\!6$ (E).}
    \label{fig:main_res}
    \vspace{-\baselineskip}
\end{figure*}

\subsection{AdaGCPI: A Meta-algorithm for Good Composite Subpopulation Identification}

\squish{To solve the GCPI problem, we propose AdaGCPI, an \textit{Ada}ptive \textit{G}ood \textit{C}omposite sub\textit{P}opulation \textit{I}dentification meta-algorithm, as formalized in Algorithm \ref{alg:gpi}. At each time step $t$, the algorithm proceeds by uniformly sampling all subgroups in the active set $\mathcal{A}_{t}$ by enroling two patients from each. For ease of presentation we assume equal sized subgroups ($\pi_j=\frac{1}{K}$) here but note that this could easily be avoided by sampling (with replacement) $K$ indices from the active set according to the subgroup prevalence $\pi_j/\sum_{i \in \mathcal{A}_{t}}\pi_i$. We then apply an identification criterion $\mathcal{I}$ that tests for evidence of an \textit{average} positive subpopulation effect across the active set. Upon success, the algorithm terminates; when evidence is not statistically significant, removal criterion $\mathcal{R}$ checks whether groups should be eliminated before enrolment continues. We discuss identification and removal criterion in turn below. }

\textbf{Identification criterion: Ensuring (approximate) FWER control.} A full Bonferroni-style adjustment would require the significance level to be adjusted by $2^K$, the number of hypotheses that could \textit{potentially} be tested. As we only select \textit{at most} $K$ hypotheses for testing in practice, this adjustment is clearly overly conservative. If the $K$ hypothesis tests were independent\footnote{To gain further intuition, let $T_{\mathcal{S}}$ denote whether hypothesis ${H}_\mathcal{S}$ is selected for testing at any time, and $R_{\mathcal{S}}$ whether it is rejected. Using an argument adapted from the discussion of discard-spending in \cite{tian2021online}, we note that $FWER \leq \mathbb{E}[\sum_{\mathcal{S}: \theta_\mathcal{S} \leq 0} T_{\mathcal{S}} R_{\mathcal{S}}]$ by Markov's inequality. Further, $\textstyle     \mathbb{E}[\sum_{\mathcal{S}: \theta_\mathcal{S} \leq 0} T_{\mathcal{S}} R_{\mathcal{S}}] = \sum_{\mathcal{S}: \theta_\mathcal{S} \leq 0}\mathbb{E}[R_{\mathcal{S}}|T_{\mathcal{S}}=1]P(T_{\mathcal{S}}=1)$. If the data used to determine hypothesis selection $T_{\mathcal{S}}$ was independent of that used to determine rejection $R_{\mathcal{S}}$, we would have that $\mathbb{E}[ R_{\mathcal{S}}|T_{\mathcal{S}}=1]= \mathbb{E}[R_{\mathcal{S}}]=\mathbb{P}(\cap^\infty_{t=1} \{\hat{\theta}_\mathcal{S} - \theta_\mathcal{S}  \geq \phi(t, \frac{\alpha}{K})\})\leq \frac{\alpha}{K}$ so that $\mathbb{E}[\sum_{\mathcal{S}: \theta_\mathcal{S} \leq 0} T_{\mathcal{S}} R_{\mathcal{S}}] \leq \frac{\alpha}{K}\mathbb{E}[\sum_{\mathcal{S}: \theta_\mathcal{S} \leq 0} T_{\mathcal{S}}] \leq \frac{\alpha}{K} K$ as at most $K$ hypotheses will be tested.}, we could use 
\begin{equation*}\mathcal{I}^K_{BF}(\mathcal{D}_t, \alpha)=\mathbbm{1}\{\hat{\theta}_{\mathcal{A}_t, N_{\mathcal{A}_t}(t)} - \phi(N_{\mathcal{A}_t}(t), \frac{\alpha}{K})>0\}
\end{equation*}
Clearly, they are not independent as datasets used for testing overlap, so identification using $\mathcal{I}^K_{BF}$ will not lead to exact FWER control. However, between selection and testing of a new hypothesis, at least $|\mathcal{A}_t|$ new samples accrue (and often many more), so any dependence decreases due to the online data collection. In experiments (Appendix D), we observe that FWER-$\alpha$ seems to hold empirically when using $\mathcal{I}^K_{BF}$, so we rely on it in our implementations. 

\textbf{Removal criterion: Exploiting subgroup \textit{and} subpopulation signals.} Using criterion $\mathcal{R}_{fut}(\mathcal{D}_t, \theta_{min}, \beta)$ as in AdaGGI, we remove \textit{individual} subgroups for futility if their individual effects are insufficient. In addition, we exploit full subpopulation information by realising that the event $\mathcal{F}_t=\mathbbm{1}\{\hat{\theta}_{\mathcal{A}_t, N_{\mathcal{A}_t}(t)} + \phi(N_{\mathcal{A}_t}(t), \beta)<\theta_{min}\}$ provides evidence that \textit{at least} one subgroup has no sufficient treatment effect. Thus, if $\mathcal{F}_t$ is true, we remove the empirically worst subgroup through the rule $\mathcal{R}_{pop\textrm{-}fut}(\mathcal{D}_t,\mathcal{A}_t, \theta_{min}, \beta) = \arg \min_{j \in \mathcal{A}_t} \hat{\theta}_{j, N_j(t-1)} - \phi(N_j(t-1), \alpha) \textrm{ if } \mathcal{F}_t \textrm{ else } \emptyset$.

\section{Experiments}

\subsection{Stylized Simulations: Understanding the (Dis)advantages of Different Strategies}

\textbf{Setup:} In this section, we consider a stylized simulation setup to gain insight into the (dis)advantages of different sampling strategies and algorithms. Only here we assume that we observe a treatment effect signal $Y^\theta_t \sim \mathcal{N}(\theta_{J_t}, 1)$ \textit{directly}; this also ensures that all our observations immediately generalize to the good \textit{arm} identification problem. We consider $K=10$ groups, $\textstyle \pi_j=\frac{1}{K}, \forall j \in \mathcal{K}$ and let $\theta_{min}=0.5$, $\alpha=0.05$, $\beta=0.1$. In the main results presented in Fig. \ref{fig:main_res}, we let $\theta_k \in \{\theta_b, \theta_g\}$, where $\theta_b=0$ and $\theta_g=0.5$ unless stated otherwise, and vary $n_{g}=|\{j: \theta_j\geq0.5\}|$. Throughout, we do not restrict budget and report $t_{stop}$, the stopping time of the algorithm (i.e. the time when \textit{all} subgroups are classified as good or not), as well as $t^{id,j}_{good}$ and $t^{id, j}_{bad}$, the time taken to identify the $j^{th}$ good group and to discard the $j^{th}$ bad group, respectively; doing so allows us to understand what the algorithm would have identified given \textit{any} budget. We compare AdaGGI with different sampling strategies -- $\mathcal{E}_{UCB}$, $\mathcal{E}_{LCB}$ and $\mathcal{E}_{LUCB}$ as discussed in Sec. \ref{sec:adaggi}, as well as two baselines (discussed further in Appendix \ref{app:bandits}): $\mathcal{E}_{unif}$, which uniformly samples groups that have not yet been identified, and $\mathcal{E}_{APT}$, which corresponds to \cite{locatelli2016optimal}'s thresholding bandit solution -- to AdaGCPI with different removal strategies ($\mathcal{R}_{fut}$ and $\mathcal{R}_{fut} + \mathcal{R}_{pop\textrm{-}fut}$). Some existing bandit algorithms arise as special cases of AdaGGI for the various sampling strategies we consider (see Appendix A.2.1). We discuss insights in turn below and present additional results in Appendix \ref{app:results}.

\textbf{Natural stopping times.} In Fig. \ref{fig:main_res}A, we investigate \textit{how long} it would take the different algorithms to select/discard \textit{all} subgroups (arms) for different $n_{g}$. First, we observe that the sampling strategy of AdaGGI has no impact on the stopping time; this is expected as identification of the final/worst group determines $t_{stop}$. Second, the total time to termination \textit{increases} as $n_{g}$ increases for AdaGGI because the identification criterion is \textit{stricter} than the removal criterion.  Third, AdaGCPI($\mathcal{R}_{fut}$), which is identical to AdaGGI($\mathcal{E}_{unif}$) except for the subpopulation-based identification criterion, performs identically to AdaGGI when $n_{g}\leq 1$ but begins to terminate earlier when $n_g$ increases as sample size can be shared across $n_{g}\geq 2$ good subgroups. Finally, AdaGCPI($\mathcal{R}_{fut}+\mathcal{R}_{pop\textrm{-}fut}$) terminates fastest throughout, as it shares statistical strength across subgroups \textit{both} when discarding and accepting subgroups; thus, the more homogeneous the population ($n_{g}$ close to 0 or 10) the faster it terminates.

\textbf{Time to identify the $j^{th}$ good group.} In Fig. \ref{fig:main_res}B\&C, we investigate \textit{when} the different algorithms make \textit{good} group discoveries, for $n_{g}=4, 8$. When comparing algorithms, we find that AdaGGI generally makes the \textit{first} discovery before AdaGCPI, as AdaGCPI makes \textit{all} discoveries at the same time (yet this often happens before AdaGGI even makes its second discovery). When comparing sampling strategies within AdaGGI, major differences become visible. (Non-adaptive) uniform sampling now clearly appears suboptimal; as expected, the thresholding approach $\mathcal{E}_{APT}$, focussing on the groups hardest to distinguish from the threshold, performs even worse. Within the other adaptive strategies, $\mathcal{E}_{LCB}$ indeed makes the first discoveries faster than $\mathcal{E}_{UCB}$ in this setting, as the latter will unnecessarily switch between good groups as upper bounds cross (because the underlying means are identical); as expected, $\mathcal{E}_{LUCB}$ lies inbetween.

\begin{figure}[b]
    \centering
    \includegraphics[width=0.9\columnwidth]{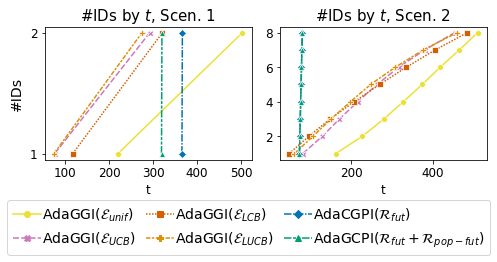}
      \vskip -0.17in
    \caption{Good group identifications over time for two additional scenarios.}
    \label{fig:diffmean}
\end{figure}

\squish{If the good groups were to exhibit quantitatively very different effects, the group with the largest $\theta_j$ should need least samples to be discovered -- thus we would expect UCB-type strategies that haven proven successful in \textit{best arm} identification \cite{jamieson2014best} to be advantageous in this context. In Fig \ref{fig:diffmean}, we therefore further investigate the relative performance of sampling strategies when altering the underlying simulation: when the means in good groups are very different (Scen. 1: $\theta_1\!=\!0.5, \theta_2\!=\!1$; $\theta_j=0, j>2$) the relative performance indeed reverses. With more good arms and less spacing between means (Scen. 2: $\theta_j\!=\!0.5+ \frac{0.5}{7}(j\!-\!1), j\leq 8$; $\theta_j\!=\!0, j >8$), this difference becomes less pronounced. In Appendix D, we additionally investigate how sampling strategies compare when outcome variance is known to differ across groups, and find that $\mathcal{E}_{LCB}$ can dominate as it intrinsically makes use of the fact that arms with \textit{lower} variance need less samples to be identified, while $\mathcal{E}_{UCB}$ may erroneously focus on groups with high variance.}

\textbf{Time to discard the $j^{th}$ bad group.} In Fig. \ref{fig:main_res}D\&E, we investigate when the different algorithms \textit{discard} groups that do not appear good. First, we observe that, unsurprisingly, AdaGCPI -- an algorithm operating by successive elimination -- discards groups much faster than AdaGGI (with the exception of AdaGGI($\mathcal{E}_{APT}$), which essentially acts like a more aggressive elimination algorithm due to its focus on the threshold). Second, we observe that AdaGCPI($\mathcal{R}_{fut}+\mathcal{R}_{pop\textrm{-}fut}$) indeed benefits from the population-based elimination criterion as groups are discarded faster esp. when $n_g$ small, which is when the population-based removal criterion will be met earlier. Third, we note that uniform sampling leads to faster elimination than (L)UCB-based sampling, which is expected as the latter actively avoid sampling groups that appear bad. Perhaps more surprisingly, LCB sampling leads to similarly fast discarding of the first bad groups, which we attribute to LCB being more likely to continue sampling from a group that has already been sampled often.

\begin{figure}
    \centering
    \includegraphics[width=0.9\columnwidth]{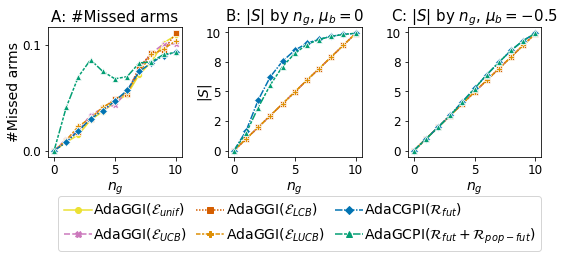}
      \vskip -0.17in
    \caption{(A): Avg. number of missed groups by $n_g$. (B\! \& \! C): Avg. |S| by $n_g$, for $\theta_b\!=0, -0.5$.}
    \label{fig:missedgroups}
    \vskip -0.2in
\end{figure}
\textbf{Incorrectly classified groups.} Finally, we consider whether subgroups are (in)correctly classified as good. First, we note that, as we show in Appendix D, Type I error is not only controlled at level $\alpha$ but essentially 0 throughout (even when we remove the Bonferroni correction); we attribute this to the used anytime confidence intervals being unnecessarily conservative as $t \!<\!\!\!< \!\infty$ here. Second, in Fig. \ref{fig:missedgroups}A, we observe that good groups are seldomly missed by either algorithm (again, likely due to conservativeness of the bounds, the rate lies far below $\beta *n_{g}$); only AdaGCPI occasionally removes a good group with the aggressive removal criterion $\mathcal{R}_{pop\textrm{-}fut}$. Third, in Fig. \ref{fig:missedgroups}B, we observe interesting differences in groups without effect that are included in the selected subpopulation $\mathcal{S}$ (note: for AdaGCPI, this does \textit{not} necessarily constitute a Type I error as long as $\theta_\mathcal{S}\!\!>\!0$). As AdaGGI identifies groups individually, $|\mathcal{S}|\!\approx\! n_{g}$ throughout, while AdaGCPI allows \textit{free-riding} of groups without effect on the larger effects of other groups, i.e. $|\mathcal{S}|\!>\! n_{g}$, especially when $n_{g}$ is large, which leads to dilution of the effect on the full subpopulation but retains the average positive effect estimate. In Fig. \ref{fig:missedgroups}C we set $\theta_b \!=\! \!-0.5$ instead of $0$, and observe that this behavior ceases when groups contribute sufficiently large negative effects.

\begin{table}
\vspace{-6pt}
\centering
\caption{Results of 1000 simulated trials: Prop. of successful trials, avg. size of discovered subpopulation and, as prop. of budget: Avg. time to termination and avg. time to identification of the first good and bad group.}\label{table:textres}
\vspace{3pt}
\small
\resizebox{\linewidth}{!}{\begin{tabular}{l|llllll}\toprule
Scenario: $\bm{\theta}$          & Method  & \%Succ.                        & $|\mathcal{S}|$ & $\frac{t_{stop}}{B}$           & $\frac{t_{1g}}{B}$             & $\frac{t_{1b}}{B}$             \\ \midrule
A:$[0, 0, 0]$       & GSDS    & 2.6                            & 0.04            & 0.74                           &                                & 0.5                            \\
                  & AdaGGI  & \textbf{0}    & 0               & 0.64                           &                                & 0.24                           \\
                  & AdaGCPI & \textbf{0}    & 0               & \textbf{0.49} &                                & \textbf{0.23} \\ \midrule
B:$[\!-\!0.2, 0, 0.2]$  & GSDS    & \textbf{99.3} & 1.19            & 0.64                           & 0.64                           & 0.5                            \\
                  & AdaGGI  & 97.9                           & 0.98            & 0.63                           & \textbf{0.46} & 0.38                           \\
                  & AdaGCPI & 95                             & 1.04            & \textbf{0.61} & 0.61                           & \textbf{0.15} \\\midrule
C:$[0, 0.1, 0.3]$   & GSDS    & \textbf{100}  & 2.03            & \textbf{0.50} & 0.50                           & 0.50                           \\
                  & AdaGGI  & 99                             & 1.00            & 0.55                           & \textbf{0.29} & 0.59                           \\
                  & AdaGCPI & 89                             & 2.28            & 0.89                           & 0.55                           & \textbf{0.44} \\ \midrule
D:$[0.2, 0.2, 0.2]$ & GSDS    & \textbf{100}  & 2.98            & 0.50                           & 0.5                            &                                \\
                  & AdaGGI  & 99.8                           & 2.27            & 0.94                           & \textbf{0.36} &                                \\
                  & AdaGCPI & 99.8                           & 2.99            & \textbf{0.37} & 0.37                           &                                \\ \midrule
E:$[0.3, 0.3, 0.3]$ & GSDS    & 100                            & 3               & 0.5                            & 0.5                            &                                \\
                  & AdaGGI  & 100                            & 3               & 0.49                           & \textbf{0.16} &                                \\
                  & AdaGCPI & 100                            & 3               & \textbf{0.17} & 0.17                           & \\   \bottomrule
\end{tabular}}
\vspace{-\baselineskip}
\vspace{-6pt}
\end{table}

\subsection{Application: Simulating a Clinical Trial} 

\squish{Finally, we apply our methods to a clinical trial setup. Because ground truth treatment effects are never observed in real data, papers on adaptive clinical trials (and the literature on treatment effect estimation more generally \cite{curth2021really}) usually have to resort to simulation studies to evaluate the quality of their algorithms (e.g  \cite{magnusson2013group, atan2019sequential}), where simulations are often semi-synthetic in that they are designed to reflect some qualities of real data. Here, we do so by building off the simulation setting presented in Section 6 of \cite{magnusson2013group}, which is
in turn motivated by the I-SPY 2 breast cancer trial for neoadjuvant therapies \cite{barker2009spy}.  We consider 3 equal sized subgroups with unknown treatment effect vector $\bm{\theta}=[\theta_1, \theta_2, \theta_3]$ and as \cite{magnusson2013group} let $\theta_{min}\!=\!0.2, \alpha\!=\!0.025$ and $\beta\!=\!0.1$. Their setup considers binary outcomes ($Y^C_j \sim \mathcal{B}(\mu_{0,j}), Y^T_j \sim \mathcal{B}(\mu_{0,j}+\theta_j)$); in Appendix D we also consider normal outcomes. Using their budget calculations we set a budget of $B=800$ pairs of patients. We compare AdaGCPI and AdaGGI to \cite{magnusson2013group}`s proposed GSDS procedure as a baseline, which is structured similarly to AdaGCPI but (i) allows only $n_a$ (preprespecified) interim analyses (in their study and here $n_a\!\!=\!2$, allowing a single interim analysis halfway), (ii) selects and fixes subpopulation $\mathcal{S}$ at the first interim analysis and (iii) relies on explicitly calculated normal error-spending boundaries. GSDS and the simulation are further described in Appendix \ref{app:details}.}

 \squish{The original experiment in \cite{magnusson2013group} has $\bm{\theta} \approx [0, 0.05, 0.1]$, i.e. all $\theta_j\! <\! \theta_{min}$, so that none of the designs are powered to detect any effect; indeed we find that across 1000 replications GSDS declares the trial successful $67\%$ of the time, while AdaGGI and AdaGCPI\footnote{{We focus on comparison with GSDS and use Sec. 5.1's overall best versions, AdaGGI($\mathcal{E}_{LCB}$) and AdaGCPI($\mathcal{R}_{fut}\!+\!\mathcal{R}_{pop\textrm{-}fut}$),  as AdaGGI and AdaGCPI; full results are in Appendix \ref{app:results}.}} declare success only in $13\%$ and $7\%$ -- a direct consequence of our designs discarding effects below the minumum clinically relevant $\theta_{min}$. To gain more interesting insights into relative performance, we therefore consider five scenarios with varying $\bm{\theta}$ in Table \ref{table:textres}.
We observe that GSDS generally has more power to detect smaller effects. This is not surprising because (i) GSDS does not automatically discard groups below $\theta_{min}$ and (ii) the used anytime confidence intervals in both our algorithms are, as discussed above, overly conservative -- especially when compared to the exact normal confidence bounds used in GSDS. Nonetheless, compared to our fully adaptive approaches, GSDS suffers from its rigidity (i.e. being restricted to pre-specified interim analysis points). In Scenarios B-D, it is apparent that both AdaGGI and AdaGCPI can make judgements about a single subgroup much before GSDS' first scheduled interim analysis (as before, AdaGGI generally finds the first good group faster, while AdaGCPI discards the first bad subgroup faster). In Scenarios A\&E, where outcomes are extreme (all $\theta_j\!=\!0$ and $\theta_j\!>\!\theta_{min}$, respectively), the advantage of the flexibility of AdaGCPI relative to GSDS is most obvious, as, due to the lack of restriction on analysis points, AdaGCPI can terminate \textit{much} earlier than the first scheduled interim analysis of GSDS.}

{In summary, we thus find that our algorithms can outperform GSDS in some scenarios because they are much less constrained in terms of \textit{when} they can terminate. At the same time, their performance in other scenarios is limited by the used anytime confidence intervals  $\phi(\cdot, \cdot)$, which are less tight than the exact normal intervals used in GSDS. Investigating the use of other confidence intervals to instantiate our meta-algorithms would thus be an interesting avenue for future work.}

\section{Conclusion}\label{Sec:concl}

\squish{We investigated how to adaptively identify patient subpopulations with treatment benefit during a clinical trial using ideas from ML, and proposed two problem formulations and associated meta-algorithms with different characteristics. We highlighted that the elimination-based AdaGCPI algorithm generally terminates using fewer samples, but may include subgroups that have no true benefit from treatment in the selected subpopulation if other groups have a sufficiently positive effect. Using AdaGGI, which discovers individual subgroups, this can generally be avoided -- if one is willing to use substantially more samples. As we discuss further in Appendix \ref{app:ext}, we believe that the formalization of the population identification problem presented in this paper opens up many interesting avenues for future ML research in this context. In particular, we believe there is great potential for extending our setting to incorporate further practical requirements -- e.g. allowing for delayed feedback or discovery of (not pre-specified) subgroups -- and theoretical analysis of the considered algorithms and sampling strategies.}

\textbf{Societal Impact.} There is a clear (ethical) tradeoff when deciding between algorithms to use in practice: AdaGCPI has the advantage that it may allow to bring a novel treatments to larger audiences faster and, due to uniform enrolment, does not give (arbitrary) preference to a single subgroup -- but it may lead to prescription recommendations that include subgroups without effect. Conversely, AdaGGI has the advantage that it will recommend treatment only in truly good subgroups, yet highly non-uniform enrolment may lead to fairness concerns (e.g. due to the randomness in deciding which equally good group to recruit first) and trials may require much larger sample sizes and hence delay the release of a potentially life-saving treatment.  \cite{freidlin2013phase} discusses similar issues for enrichment designs more generally. 

Further, we note that the choice of using of adaptive trial designs instead of conventional non-adaptive trials is always highly situational \cite{palmer1999ethics}. There are definitely cases where one may want to continue to rely on conventional trial designs – e.g. applications where there are very large delays between patient enrolment and realisation of outcomes. There are, however, also cases in which adaptivity can be expected to be beneficial (it is sometimes even argued that adaptive designs are the only ethically permissible experimental designs, see e.g. \cite{fillion2019clinical} for a discussion). We therefore believe that adaptive enrichment designs like ours would be of most value in applications where high between-subgroup variability of effectiveness is expected.

\section*{Acknowledgements} We would like to thank anonymous reviewers for insightful comments and discussions on earlier
drafts of this paper. AC gratefully acknowledges funding from AstraZeneca. AH is funded by the US Office of Naval Research (ONR).

\bibliographystyle{unsrt}
\bibliography{main}


\newpage
\appendix
\onecolumn
\section{Appendix A: Additional Literature Review}

\subsection{Extended review of adaptive clinical trial literature on enrichment designs}\label{app:addtrial}
Below, we discuss in some more detail the clinical trials literature on adaptive enrichment trials which allow discontinuation of subgroups and changes of the population (and hence hypothesis) under consideration in a clinical trial. We focus here on designs where the subgroups under investigation are prespecified; subgroup discovery in the presence of single or multiple biomarkers is covered in e.g. the literature on so-called \textit{adaptive signature designs}. \cite{freidlin2005adaptive, freidlin2010cross, zhang2017subgroup, simon2013adaptive}. For a broader review of adaptive enrichment designs, refer to \cite{thall2021adaptive}.

\cite{jennison2007adaptive} (Section 6) describe a generic two-stage enrichment design with $K$ subpopulations that are \textit{not} necessarily disjoint (as in our case) or nested, which allows for selection of an arbitrary population $j^*$ after the first stage, after which recruitment is focussed on $j^*$ and the hypothesis to be tested is $H_{0j*}$, where error is controled through application of closed testing procedures \cite{henning2015closed}. \cite{wang2009adaptive} also consider two-stage trials under different types of restrictions with multiple nested subpopulations determined by biomarker interactions. 

\cite{jenkins2011adaptive, friede2012conditional, brannath2009confirmatory} all consider a setting where one can either consider the full population or a \textit{single} pre-specified subpopulation of heightened interest; at an interim analysis it is to be decided whether to continue with the full population or within the subpopulation only (or not at all). These designs differ in both the rules for population selection and the hypothesis tests used, but all rely on closed testing principles. \cite{stallard2014adaptive} empirically compares some of these and other approaches for population selection and hypothesis testing in the adaptive enrichment problem with two subpopulation and a single interim analyses. 

\cite{rosenblum2016group} also considers a setting where either the full population or a single promising subgroup is of interest, however, instead of only one interim analysis the trial has multiple analyses where the trial can be stopped for efficacy/futility in either the full population or the subgroups based on normal stopping boundaries. Finally, \cite{magnusson2013group} propose a design  (\cite{chiu2018design} discuss a multistage design analogous to theirs) that is most closely related to our AdaGCPI approach, where the main differences  lie in that (i) \cite{magnusson2013group} \textit{fix} the selected subpopulation after the first stage and (ii) exact probability boundaries are calculated for termination. We describe \cite{magnusson2013group}'s proposed GSDS procedure in more detail in Appendix C. 

The most relevant related work from the ML literature that we are aware of is \cite{atan2019sequential}; they also consider adaptive recruitment to discover all good subgroups and do so using a Bayesian MDP-based design that learns by optimizing an objective function that \textit{trades off} Type I and II error given a limited budget. As such, type I error is neither controlled nor is multiplicity considered, making this approach (objective) conceptually, i.e. abstracting away specific implementation choices, most similar to good arm identification and thresholding bandits under a \textit{fixed budget} (only) setting. Finally, \cite{huyukmake} recently considered subpopulation selection in adaptive clinical trials but for \textit{portfolio-level management} of trials rather than the sample-by-sample decisions we consider here.

\subsection{Extended contextualization of the GGI and GCPI problems within bandit literature}\label{app:bandits}

GGI and GCPI are closely related to multi-armed bandit problems as one can interpret each considered subgroup as an \textit{arm} and their (unknown) treatment effect as the \textit{mean reward} of that arm. Typically, the goal in a bandit problem is to maximize the rewards of all arms that are ``played''  (e.g.\ \cite{auer2002finite}). Since the mean rewards are unknown initially, this requires striking a balance between \textit{exploring} arms to gain information about their rewards and \textit{exploiting} arms that appear to have high rewards. In our setting, this conventional objective would have corresponded to maximizing the benefit received by all patients recruited into the trial. Instead, we focus on what is known as \textit{pure exploration} in the bandit literature, where the rewards of played arms do not matter except for that of a singular arm identified at the end \cite{bubeck2009pure,bubeck2011pure,chen2014combinatorial,degenne2019pure}.

Different purely-exploratory objectives have been considered in the multi-armed bandit literature. Best arm identification (BAI) problems aim to identify the arm (or the top-$K$ arms) with the largest mean reward (e.g.\ \cite{audibert2010best}). Here, the success can be measured via the reward gap between the identified arm and the true best arm. In the \textit{fixed budget} setting, the goal is to maximize the probability of the identified arm indeed being the best given a fixed budget of samples \cite{gabillon2012best,bubeck2013multiple,carpentier2016tight}, while in the \textit{fixed confidence} setting, the goal is to minimize the number of samples necessary to guarantee a fixed level of confidence \cite{maron1993hoeffding,even2006action,mnih2008empirical,kalyanakrishnan2010efficient,gabillon2012best,kalyanakrishnan2012pac,jamieson2014best,garivier2016optimal}. Good arm identification (GAI) problems (sometimes called pure exploration in thresholding bandits) aim to identify arms with mean rewards that are higher than a pre-specified threshold. These problems too can be considered either in fixed budget \cite{locatelli2016optimal,mukherjee2017thresholding,katz2018feasible} or fixed confidence \cite{kano2019good,katz2020true} settings.

GGI is essentially a type of GAI problem but it requires both the budget as well as the confidence in each identified arm being good to be fixed, and given those constraints, aims to identify as many good arms as possible. In existing formulations of GAI, the aim is usually to identify \textit{all} good arms, which is only possible with the more relaxed constraint of either just the budget or the confidence being fixed (but not both at the same time). GCPI is similar to GGI in that it too requires both the budget and the confidence to be fixed but it only aims to identify a collection of arms that are good \textit{on average}\footnote{\cite{russac2021b} also consider a GAI problem involving weighted averages over collections of subgroups in a population, but there, the viable collections and weights are fixed and not part of the optimization problem, which is very different from the GCPI problem where the collections are optimized over.} rather than arms that are all individually good. Table~\ref{tab:related-bandit} formally compares GGI and GCPI with the existing pure exploration problems.

\begin{table}
    \caption{Comparison of pure exploration problems. GGI and 
    GCPI uniquely require both the budget as well as the confidence to be fixed, and aim to identify as many suitable arms as possible within those constraints. In contrast, other problems aim to identify all suitable arms, which is only possible with the more relaxed constraint of either just the budget or just the confidence being fixed. FB and FC stand for fixed budget and fixed confidence respectively.}
    \label{tab:related-bandit}
    \resizebox{\linewidth}{!}{
        \small
        \begin{tabular}{@{}l*5{c}l@{}}
            \toprule
            \bf Problem & \bf Ref.\ & \bf\makecell{Type of\\arms identified} & \bf\makecell{Number of\\arms identified} & \bf Budget & \bf Confidence & \bf Formulation \\
            \midrule
            BAI & \cite{audibert2010best} & \multirow{3}{*}{\makecell{Best arms\\ $i^*=\argmax_i\theta_i$}} & \multirow{3}{*}{Top-$K$ arms} & Variable & Variable & minimize~~$\theta_{i^*}-\theta_{\hat{\imath}^*}$ \\
            BAI w/ FB & \cite{carpentier2016tight} & & & Fixed ($T$) & Maximized & maximize~~$\mathbb{P}(\hat{\imath}^*(T)=i^*)$ \\
            BAI w/ FC & \cite{garivier2016optimal} & & & Minimized & Fixed ($1-\delta$) & minimize~~$T$~~s.t.~~$\mathbb{P}(\hat{\imath}^*(T)\neq i^*)\leq\delta$ \\
            \midrule
            GAI w/ FB & \cite{locatelli2016optimal} & \multirow{2}{*}{\makecell{Good arms\\$\mathcal{I}=\{i:\theta_i>\xi\}$}} & \multirow{2}{*}{All good arms} &  Fixed ($T$) & Maximized & maximize~~$\mathbb{P}(\hat{\mathcal{I}}(T)=\mathcal{I})$ \\
            GAI w/ FC & \cite{katz2020true} & & & Minimized & Fixed ($1-\delta$) & minimize~~$T$~~s.t.~~$\mathbb{P}(\hat{\mathcal{I}}(T)\neq\mathcal{I})\leq\delta$ \\
            \midrule
            \bf GGI & \bf (Ours) & \makecell{Good arms\\$\mathcal{I}=\{i:\theta_i>\xi\}$} & Maximized & Fixed ($T$) & \makecell{Fixed ($1-\delta$)\\w.r.t.\ type I error} & maximize~~$|\hat{\mathcal{I}}(T)|$~~s.t.~~$\mathbb{P}(\hat{\mathcal{I}}(T)\setminus\mathcal{I}\neq\emptyset)\leq\delta$ \\[9pt]
            \bf GCPI & \bf (Ours) & \makecell{Good composite arms\\$\mathcal{I}:\frac{1}{|\mathcal{I}|}\sum_{i\in\mathcal{I}}\theta_i>\xi$} & Maximized & Fixed ($T$) & Fixed ($1-\delta$) & maximize~~$|\hat{\mathcal{I}}(T)|$~~s.t.~~$\mathbb{P}\Big(\frac{1}{|\hat{\mathcal{I}}(T)|}\sum_{i\in\hat{\mathcal{I}}(T)}\theta_i\not>\xi\Big)\leq\delta$ \\
            \bottomrule
        \end{tabular}}
\end{table}

\subsubsection{How existing pure exploration solutions arise as special cases of AdaGGI} 
One of the main goals of this paper is to formalize, contextualize and understand the  trial population identification problem as a pure exploration bandit problem. Because our paper considers a new problem formulation, there -- to the best of our knowledge -- are no off-the-shelf solutions from the bandit literature that have already solved this exact problem. Therefore, this paper studies how to apply and adapt solutions proposed for related problems and empirically investigates how different approaches work in our context.

To do so, we study very generic meta-algorithms, which give rise to adaptations of some existing combinatorial bandit solutions as special cases, allowing for fair comparison of different approaches. Note that both the thresholding bandit and good arm identification (GAI) are combinatorial bandit instances and their specific problem formulations are closer to our problem setting than generic combinatorial bandits, making their solutions more likely to perform well in our context. Below, we discuss in detail how GAI algorithms, thresholding bandits and a generic combinatorial bandit solution arise as variations of AdaGGI and can thus be seen as `bandit baselines' in our experiments. 

\paragraph{GAI algorithms -- AdaGGI($\mathcal{E}_{UCB}$)} As outlined in Section 3, the GAI algorithms proposed in \cite{kano2019good, jamieson2018bandit} proved most suitable to adapt to our setting and thus share a very similar backbone to AdaGGI. The main conceptual differences to existing implementations lies in that (i) they exclusively rely on UCB-sampling and (ii) have no \cite{jamieson2018bandit} or a stricter \cite{kano2019good} removal criterion.  The special case $\mathcal{E}_{UCB}$ could thus be seen as a GAI-bandit baseline with adapted removal criterion. Adaptation of the removal criterion to allow discarding of groups without clinically relevant effect greatly improves those algorithms with respect to stopping time; the original criteria lead to infinite running times when `bad’ group effects are exactly zero (as in our experiments).

\paragraph{Thresholding bandit -- AdaGGI($\mathcal{E}_{APT}$)} Another approach that could be adapted to our setting is \cite{locatelli2016optimal}'s thresholding bandit solution. Because the thresholding bandit problem aims at correctly classifying \textit{all} arms as either good or bad using a fixed budget, \cite{locatelli2016optimal}'s Anytime Parameter-free Thresholding (APT) algorithm tries to equalize the confidence in the classification of all arms by ensuring that $N_j(t)(\hat{\theta}_{j, N_j(t)}-\theta_0)^2$ is constant across arms. This corresponds to a sampling strategy $\mathcal{E}_{APT}(\mathcal{D}_{t-1}, \mathcal{A}_{t-1}) = \arg \min_{j\in \mathcal{A}_{t-1}} \sqrt{N_j(t-1)} (\hat{\theta}_{j, N_j(t-1)}-\theta_0)$ with $\theta_0=0$ in our setup. Conceptually, this will lead to sampling the groups that are \textit{furthest} from being identified -- this is the opposite strategy to what $\mathcal{E}_{LCB}$ tries to accomplish and cannot be expected to perform well in our context. Because the original paper \cite{locatelli2016optimal} focusses on a fixed budget only setting, it is lacking some form of identification and removal criterion. We therefore instantiate it using the AdaGGI backbone and simply use $\mathcal{E}_{APT}$ as the sampling strategy. 

\paragraph{Generic combinatorial bandit baseline -- AdaGGI($\mathcal{E}_{unif}$)} Finally, we consider adapting more generic combinatorial bandit solutions, which generally aim to optimize some objective over collections of arms. Here, we consider \cite{chen2013combinatorial}'s Combinatorial Upper Confidence Bound (CUCB) algorithm  in more detail as it permits straightforward adaptation to our setting. The general setting considered in \cite{chen2013combinatorial} allows to play a super-arm $\mathcal{S}$ at each time $t$, and their algorithm assumes existence of an oracle that outputs the optimal $\mathcal{S}$ whenever provided with the underlying distributions of all arms; in the GAI setting this simply picks all arms whose means exceed the threshold. The algorithm proceeds by constructing upper confidence bounds $\tilde{\theta}_{j, t}=\hat{\theta}_{j, N_j(t-1)} + \phi(N_j(t-1), \beta)$ on the means of all arms, and then applies the oracle to the $\tilde{\theta}_{j, t}$, outputting a super-arm $\mathcal{S}_t$ to sample. In our context, this would sample all arms for which it holds that $\hat{\theta}_{j, N_j(t-1)} + \phi(N_j(t-1), \beta) > \theta_0$. Note that this essentially corresponds to AdaGGI with removal criterion $\mathcal{R}_{fut}(\mathcal{D}_t, \theta_{0}, \beta)$ instead of $\mathcal{R}_{fut}(\mathcal{D}_t, \theta_{min}, \beta)$, and uniform sampling of the active set. As discussed above, setting $\theta_{min} \neq \theta_0$ can only improve the algorithm’s performance; thus AdaGGI($\mathcal{E}_{unif}$) -- i.e. simple uniform sampling of the active set -- corresponds to a straightforward adaptation of the CUCB algorithm to our setting.

\section{Appendix B: Possible Extensions and Future Work}\label{app:ext}
 We believe that this paper opens up many interesting avenues for future research; natural next steps lie in (i) extending the setting under consideration to incorporate more realistic problem features and (ii) further studying and improving components of the algorithms. 

\textbf{Extending the setting.} Multiple modifications to the data generating process might lead to a more realistic setting and interesting research problems at the same time:
\begin{itemize}
    \item \textbf{Considering batched (grouped) observations: } In practice, it might be operationally difficult to collect and reveal \textit{individual} patient responses as they come in; instead it might be more easily feasible to release patient responses in \textit{batches} or \textit{groups} as is commonly done in \textit{group sequential designs} \cite{pocock1977group}. AdaGCPI could directly accommodate this: instead of recruiting $|\mathcal{A}_t|$ patient pairs uniformly and evaluating the subpopulation immediately, a larger batch of patients could be recruited (uniformly from the active set) before using the updated dataset for testing the hypothesis. Doing the same for AdaGGI may not be optimal, as -- because the original sampling strategies are \textit{deterministic} -- one would then have to recruit an entire batch of patients from the same subgroup, which may explore insufficiently. Instead, sampling strategies that resemble Thompson sampling \cite{thompson1933likelihood, russo2018tutorial} -- i.e. strategies that are \textit{random} and recruit patients proportionally to \textit{the probability of their subgroup being good} -- may be more suited to this scenario.
    \item \textbf{Allowing delayed feedback: } Another difficulty likely to be encountered in practice, particularly when considering time-to-event data or other long term outcomes, might be that not all outcomes of previously recruited patients are available when making the next recruitment decision. The biostatistics literature has investigated how one can use available \textit{short term outcomes} that are indicative of the long term outcomes in such scenarios \cite{huang2009using}, while the bandit literature has developed approaches for decision making under delayed feedback \cite{grover2018best}; it would be interesting to investigate how to incorporate either into our framework.
  \item \textbf{Incorporating covariates and discovering subgroups: } An interesting extension to the setting considered here would be to make use of any other patient information (context) that may be available, e.g. \textit{prognostic} information that may explain some baseline variation likely to exist in practice and hence improve precision of estimators (as in e.g. \cite{gera2020blinded}). When no subgroups are pre-specified, one may also make use of such information to \textit{discover} subgroups that differ in their treatment response through so-called \textit{adaptive signature designs} \cite{freidlin2005adaptive, freidlin2010cross, zhang2017subgroup, simon2013adaptive}; investigating how to better use ML tools to efficiently discover such subgroups may be a natural next step.
\end{itemize}

\textbf{Analyzing problem settings and algorithms.} We believe that a number of our empirical findings both motivate further theoretical analyses \textit{and} suggest that improvements to our implementations may be possible:
\begin{itemize}
    \item \textbf{Comparing problem complexity of GGI and GCPI theoretically:} Our experiments confirmed the intuition that the GCPI problem can be easier (faster) to solve than the GGI problem, especially when subgroups are close to homogeneously \textit{all} good or bad. It would be an interesting avenue for future work to confirm and analyze this theoretically.
    \item \textbf{Comparing sampling strategies theoretically:} Our experiments also confirmed the intuition that, depending on the underlying problem structure, different (non-uniform) sampling strategies are better at discovering (the first) good arm fast, and it would thus be interesting to formally derive scenarios in which either UCB or LCB strategies could be expected to have an advantage.  
    \item \textbf{Improving the used confidence intervals: } We observed in our experiments that the $\phi(\cdot, \cdot)$ that we used seemed to create overly conservative confidence intervals in our settings. One possibility to improve this may be to rely on the fact that usually $B <\!\!\!<\infty$ and to therefore construct alternatives that instead of allowing for infinitely many peeks at the data, allow only $L\leq B$ decision points which may lead to less necessity to be conservative. 
  \item \textbf{Explicitly incorporating budget in sampling and elimination strategies: } Finally, we note that it may be an interesting avenue for future work to develop a removal criterion $\mathcal{R}_{Budget}$ that forces early removal of a subgroup, either permanently from AdaGCPI whenever it is expected that there is insufficient budget left  to prove treatment effectiveness with confidence $\alpha$ in the current subpopulation (this would be the case if the average effect in the current subpopulation is likely too low to do so; more aggressively removing the subgroups that appear worst may be appropriate in this context) or temporarily from consideration for sampling in AdaGGI. Alternatively, one could also investigate new multi-stage meta-algorithms with an initial more exploratory stage and a later more exploitative stage, where the number of samples allocated to each stage or the transition between stages would depend on budget.
\end{itemize}

\section{Appendix C: Experimental Details}\label{app:details}
\subsection{Stylized simulations (Section 5.1)}
All data was generated according to the setup described in Section 5.1: There are $K=10$ groups, $\textstyle \pi_j=\frac{1}{K}, \forall j \in \mathcal{K}$ and outcomes are normally distributed according to $\mathcal{N}(\theta_k, 1)$, where $\sigma^2=1$ is assumed \textit{known}. In the main results presented in Fig. 2, we let $\theta_k=0.5$ for $k\leq n_g$ and $\theta_k=0$ for $k>n_g$, for $n_g \in \{0, \ldots, 10\}$. In Fig. 4, we set bad means equal to $-0.5$, and in Fig. 3 we let good means vary between $0.5$ and $1$ by setting them equal to $\theta_j\!=\!0.5+ \frac{0.5}{n_g-1}(j\!-\!1), j\leq n_g$. 

For all algorithms we set $\theta_{min}=0.5$, $\alpha=0.05$, $\beta=0.1$ and $n_0=1$. As $Y^\theta$ is assumed normally distributed with known variance $\sigma^2=1$, we use $\phi(t, \delta)=\sqrt{2\frac{ \!\log(1/\delta) \! +  \!3 \log \log(1/\delta) \!+ \!(3/2)\log \log(et/2))}{t}}$ as in \cite{jamieson2018bandit}. Note that we can use this for both AdaGGI and AdaGCPI as all outcomes in any subpopulation are distributed equally under the null hypothesis (regardless of subgroup, under the null hypothesis all outcomes are distributed according to $\mathcal{N}(0,1)$).

\subsection{Simulated trials (Section 5.2)}
In Section 5.2, we use a modified version of the experiment in section 6 of \cite{magnusson2013group}, which is in turn motivated by the I-SPY 2 breast cancer trial for neoadjuvant therapies \cite{barker2009spy}. The assumed end point of interest is the occurrence of pathologic complete response (pCR), \cite{magnusson2013group} assume this to follow a Bernoulli distribution where for the controls $Y^C\sim \mathcal{B}(0.4)$ for all subgroups while  the outcomes in treated individuals can differ across subgroups as $Y^T_j \sim \mathcal{B}(0.4+\theta_j)$.  As \cite{magnusson2013group} we consider 3 subgroups, for simplicity we assume them to be equal sized ($\pi_k=\frac{1}{3}$) here. In addition to the Bernoulli setting from the main text, we also consider an additional setting with normally distributed outcomes in Appendix D (with known $\sigma^2=1$) i.e.  $Y^C_j \sim \mathcal{N}(0, 1), Y^T_j \sim \mathcal{N}(\theta_j, 1), \forall j\in[3]$. 

As \cite{magnusson2013group} we let $\theta_{min}\!=\!0.2, \alpha\!=\!0.025$ and $\beta\!=\!0.1$. For our algorithms we additionally let $n_0=5$ due to the higher variance induced by considering \textit{a difference} between random variables now. As the difference between two normal random variables with variance $\sigma^2$ is normal with variance $2\sigma^2$, we use $\phi(t, \delta)=2\sqrt{\frac{ \!\log(1/\delta) \! +  \!3 \log \log(1/\delta) \!+ \!(3/2)\log \log(et/2))}{t}}$ for the normal outcomes, and, as bernoulli variables are $\frac{1}{4}$ subgaussian, we use $\phi(t, \delta)=\sqrt{\frac{ \!\log(1/\delta) \! +  \!3 \log \log(1/\delta) \!+ \!(3/2)\log \log(et/2))}{t}}$ for the difference between binary outcomes.

\textbf{Description of GSDS.} We now briefly formally describe the group sequential design for subgroups (GSDS) proposed in \cite{magnusson2013group}. The design requires: a pre-specified number of interim analyses $n_a$, a test statistic $Y_j(t)$ and associated Fisher information $\mathcal{I}_j(t)$, a desired significance level $\alpha$ and power $1-\beta$. $\mathcal{a}$ is used to calculate stopping boundaries $\{(l_p, u_p)\}^{n_a}_{p=1}$ for each interim analysis. $\beta$ is used to calculate a \textit{maximum information} level $\mathcal{I}_{max}$, which is in turn used to determine the sample size. The algorithm proceeds as follows: at the first interim analysis at time $t_1$, a subpopulation is selected through exclusion of all bad subgroups: $\mathcal{S}^*=\{j \in \mathcal{K}: Y_j(t_1)\sqrt{\mathcal{I}_j(t_1)} > l_1\}$. If $Y_{\mathcal{S}^*}(t_1)\sqrt{\mathcal{I}_{\mathcal{S}^*}(t_1)} > u_1$, the trial terminates immediately for efficacy; otherwise the trial continues and at all $n_a-1$ subsequent stages, the trial is terminated for efficacy if $Y_{\mathcal{S}^*}(t_k)\sqrt{\mathcal{I}_{\mathcal{S}^*}(t_k)} > u_k$ and terminated for futility if $Y_{\mathcal{S}^*}(t_k)\sqrt{\mathcal{I}_{\mathcal{S}^*}(t_k)} < l_k$.

\textbf{Budget calculation.} We follow the example in \cite{magnusson2013group} who calculate that for a two stage trial with $\alpha=0.025$, $\beta=0.1$ and $\theta_{min}=0.2$, we have $(l_1, u_1)=(0.7962,2.7625)$ and $l_2=u_2=2.5204$ and $\mathcal{I}_{max}=1495.5$.

In their example with binary outcomes, if we let $b$ denote the number of \textit{pairs} of recruited patients\footnote{We believe there is a typo in Sec. 6 of  \cite{magnusson2013group}, so that $n$ should denote the number of \textit{pairs} of patients, and not patients. We have adapted budget calculations accordingly}, and  $\hat{p}^C, \hat{p}^T$ the observed binary proportions in each group, we have that 
\begin{equation}
    Y=\hat{p}^T - \hat{p}^C \textrm{ and } \mathcal{I} = \frac{b}{2\tilde{p}(1-\tilde{p})}
\end{equation}
where $\tilde{p}$ is the average response rate and is conservatively set to 0.5. Solving $\mathcal{I}_{max}$ for $b$ yields a (rounded) budget $B=800$ pairs of patients.

Similarly, when doing the same for normal outcomes with known variance $\sigma^2$, if we let $\hat{\mu}^C, \hat{\mu}^T$ denote the means in treated and control arm, we have
\begin{equation}
    Y=\hat{\mu}^T - \hat{\mu}^C \textrm{ and } \mathcal{I}=\frac{b}{2\sigma^2}
\end{equation}
and with $\sigma^2=1$ this yields a rounded budget of $B=3000$.

\section{Appendix D: Additional Results}\label{app:results}

\subsection{Additional simulation results (Sec 5.1)}
\subsubsection{Identifications: Complete results}
\begin{figure}[h]
    \centering
    \includegraphics[width=0.97\textwidth]{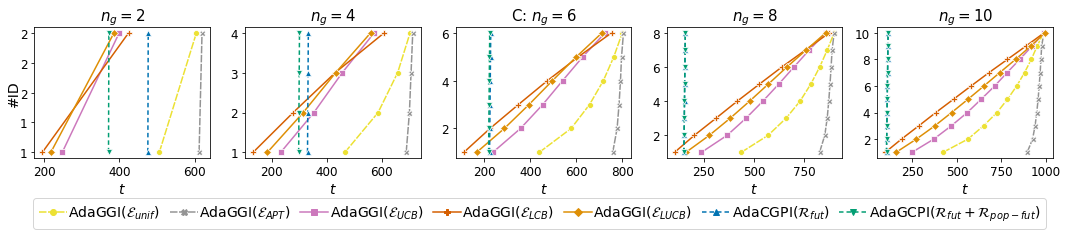}
     \vskip -0.13in
    \caption{Results describing identification of good groups over time, for $n_g \in \{2, 4, 6, 8, 10\}$; avg. across 1000 replications.}
    \label{fig:all_id}
\end{figure}

\begin{figure}[h]
    \centering
    \includegraphics[width=0.97\textwidth]{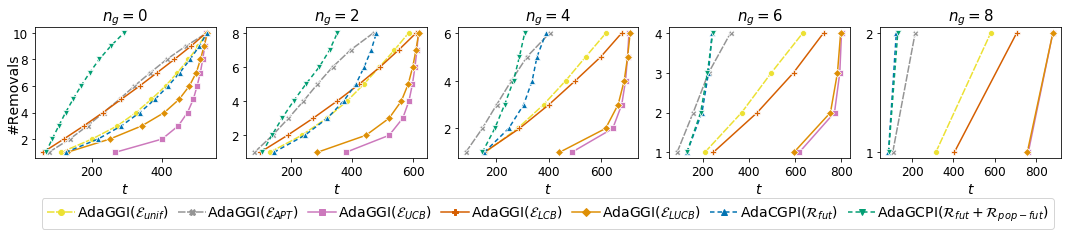}
     \vskip -0.13in
    \caption{Results describing removal of bad groups over time, for $n_g \in \{0, 2, 4, 6, 8\}$; avg. across 1000 replications.}
    \label{fig:all_rm}
\end{figure}

In Fig. \ref{fig:all_id}, we present results capturing time until identification of each good group for $n_g \in \{2, 4, 6, 8, 10\}$ (only $n_g=4, 8$ are presented in the main text). In Fig. \ref{fig:all_rm}, we present results capturing time until removal of each bad group for $n_g \in \{0, 2, 4, 6, 8\}$ (only $n_g=2, 6$ are presented in the main text). These results reflect the same insights as those presented in the main text, both in terms of comparing algorithms and in terms of comparing sampling strategies. 

\subsubsection{Type I error}
\begin{figure}[h]
    \centering
    \includegraphics[width=0.6\textwidth]{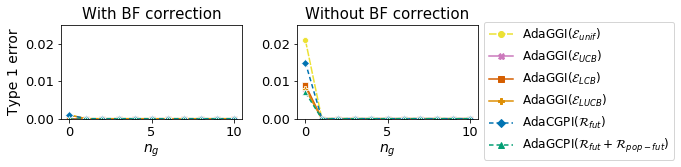}
     \vskip -0.13in
    \caption{Type 1 error (runs with \textit{any} incorrectly rejected null hypothesis) by $n_g$ across 1000 replications. Identification \textit{with} Bonferroni correction (left) and without (right).}
    \label{fig:type1}
\end{figure}

In Fig. \ref{fig:type1} we plot Type I errrors committed over 1000 simulation runs, both with and without Bonferroni correction. (Note that a Type I error is defined as \textit{any} null hypothesis being incorrectly rejected; for AdaGGI this includes any \textit{single} subgroup being incorrectly declared good, while for AdaGCPI this would mean that the selected subpopulation $\mathcal{S}$ does not have a positive \textit{average} effect.) We make multiple interesting observations: First, with Bonferroni correction, all identification criteria are clearly overly conservative -- incorrect rejections only happen when \textit{all} groups are bad, and even then this lies much below the used $\alpha=0.05$. Second, this is not primarily due to the conservativeness of the Bonferroni correction, but due to the conservativeness of the used anytime confidence interval: even when we remove the Bonferroni correction, all Type 1 errors remain below $(10-n_g)*\alpha$ (in fact, they even lie below $\alpha$). Finally, we note that the approximate Bonferroni correction we chose for AdaGCPI therefore does not appear to be problematic; in the plot \textit{without} any correction we also observe that AdaGCPI does not seem to be more likely to commit a Type I error than AdaGGI even without any correction (despite the number of hypotheses that could potentially be tested being exponential versus linear in $K$).

\subsection{Additional simulation scenarios}
\paragraph{Varying means} We present additional results on the setting presented in Fig. 3 of the main text: for $n_g \in \{2, \ldots, 10\}$ we let $\theta_j=0.5+0.5\frac{j-1}{n_g-1}$ for $j\leq n_g$ and $\theta_j=0$ otherwise. As discussed in the main text, we observe that the relative performance of sampling strategies changes in this setting: $\mathcal{E}_{LCB}$ generally performs worse than $\mathcal{E}_{UCB}$ here; with increasing $n_g$ and hence decreasing spacing between the good means, this effect reduces.
\begin{figure}[h]
    \centering
    \includegraphics[width=0.99\textwidth]{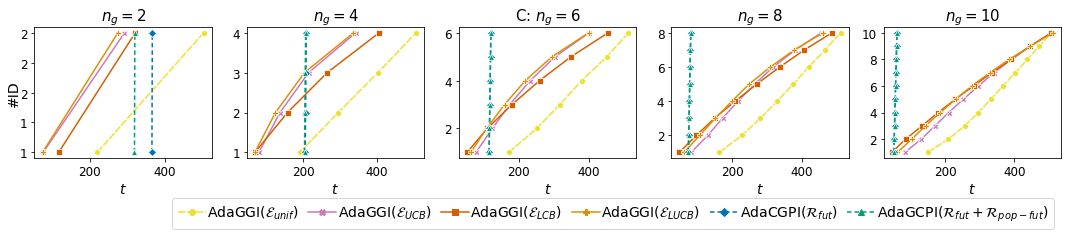}
     \vskip -0.13in
    \caption{Results describing identification of good groups over time, for $n_g \in \{2, 4, 6, 8, 10\}$ for a setting with varying means; avg. across 1000 replications.}
    \label{fig:goodidnew}
\end{figure}

\begin{figure*}[!h]
	\centering
	\subfigure[$n_g=10$]{\label{fig:varchange10}\includegraphics[width=0.45\columnwidth]{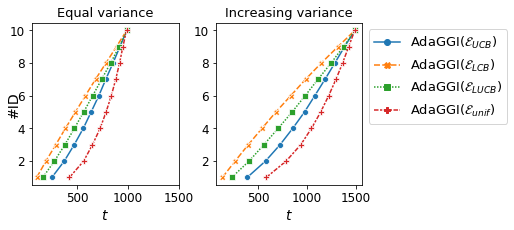}}
    \subfigure[$n_g=5$]{\label{fig:varchange5}\includegraphics[width=0.45\columnwidth]{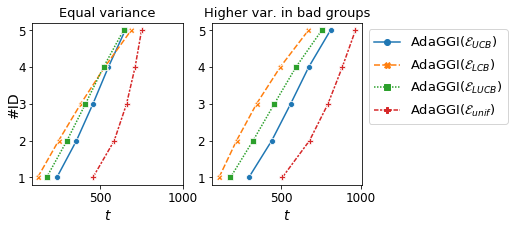}}

	\caption{Results describing identification of good groups over time, for $n_g=10$ with variances possibly increasing across groups (left) and $n_g=5$ with possibility for higher variance in bad groups (right)}\label{fig:ihdp_all}   

\end{figure*}

\paragraph{Different variances} Next we consider how changing variance affects the performance of the different sampling algorithms. In the Fig. \ref{fig:varchange10}, we compare the original setting with $n_g=10$ and $\sigma^2=1$ for all groups to one where the means are the same but $\sigma^2_j=1+\frac{j-1}{n_g}$ grows across groups. In the right Fig. \ref{fig:varchange5}, we compare the original setting with $n_g=5$ and $\sigma^2=1$ for all groups to one where $\sigma^2=2$ for the bad groups, while $\sigma^2=1$ for the good groups. We observe that identification times worsen across the board in both settings, but that the time increase for the first identifications for $\mathcal{E}_{UCB}$ is much larger than that for  $\mathcal{E}_{LCB}$ in absolute terms -- most likely because UCB-style algorithms may erroneously enrol groups with larger variance as the UCB will generally be higher for these groups.

\newpage
\subsection{Additional clinical trial simulation results (Sec 5.2)}
Finally, we present additional clinical trial simulation results, which include more versions of the two algorithms and an additional setting with normal outcomes, in Table \ref{tab:trialres-full}. We observe that the results with normal outcomes are largely in line with the results with binary outcomes. The relative performance of AdaGGI using different sampling strategies and AdaGCPI using different removal rules is also in line with what has been observed in Sec. 5.1 in the main text; in particular, $\mathcal{E}_{LCB}$ continues to dominate unless there is one group with a much better effect than others in which case $\mathcal{E}_{(L)UCB}$ works better, and the addition of $\mathcal{R}_{pop\textrm{-}fut}$ has the largest effect on AdaGCPIs performance whenever there is no effect across all groups. 
\begin{table}[h]
\setlength{\tabcolsep}{4pt}
\begin{center}
\begin{threeparttable}
	\fontsize{9}{7}\selectfont
\begin{tabular}{ll|lllll|lllll}\toprule
                  &                                                                                                          & \multicolumn{5}{l}{Binary}                                                                                                                          & \multicolumn{5}{l}{Normal}                                                                                                                          \\
$\bm{\theta}$          & Method                                                                                                   & \%Succ.                        & $|\mathcal{S}|$ & $\frac{t_{stop}}{B}$           & $\frac{t_{1g}}{B}$             & $\frac{t_{1b}}{B}$             & \%Succ.                        & $|\mathcal{S}|$ & $\frac{t_{stop}}{B}$           & $\frac{t_{1g}}{B}$             & $\frac{t_{1b}}{B}$             \\ \midrule
A:$[0, 0, 0]$       & GSDS                                                                                                     & 2.6                            & 0.04            & 0.74                           &                                & 0.5                            & 2.4                            & 0.04            & 0.75                           &                                & 0.51                           \\
                  & AdaGGI($\mathcal{E}_{LCB}$)                                                                              & {0}    & 0               & 0.64                           &                                & 0.24                           & 0                              & 0               & 0.69                           &                                & \textbf{0.25} \\
                  & AdaGGI($\mathcal{E}_{UCB}$)                                                                              & 0                              & 0               & 0.63                           &                                & 0.53                           & 0                              & 0               & 0.69                           &                                & 0.60                           \\
                  & AdaGGI($\mathcal{E}_{LUCB}$)                                                                             & 0                              & 0               & 0.64                           &                                & 0.48                           & 0                              & 0               & 0.69                           &                                & 0.54                           \\
                  & AdaGGI($\mathcal{E}_{unif}$)                                                                             & 0                              & 0               & 0.63                           &                                & 0.35                           & 0                              & 0               & 0.70                           &                                & 0.40                           \\
                  & AdaGCPI($\mathcal{R}_{fut}$)                                                                             & 0                              & 0               & 0.64                           &                                & 0.36                           & 0                              & 0               & 0.69                           &                                & 0.39                           \\
                  & \begin{tabular}[c]{@{}l@{}}AdaGCPI\\ ($\mathcal{R}_{fut} + \mathcal{R}_{pop\textrm{-}fut}$)\end{tabular} & {0}    & 0               & \textbf{0.49} &                                & \textbf{0.23} & 0                              & 0               & \textbf{0.54} &                                & 0.26                           \\ \midrule
B:$[-0.2, 0, 0.2]$  & GSDS                                                                                                     & \textbf{99.3} & 1.19            & 0.64                           & 0.64                           & 0.5                            & \textbf{97.9} & 1.18            & 0.68                           & 0.68                           & 0.5                            \\
                  & AdaGGI($\mathcal{E}_{LCB}$)                                                                              & 97.9                           & 0.98            & 0.63                           & \textbf{0.46} & 0.38                           & 96.6                           & 1               & 0.69                           & 0.52                           & 0.57                           \\
                  & AdaGGI($\mathcal{E}_{UCB}$)                                                                              & 98                             & 0.98            & 0.63                           & 0.48                           & 0.51                           & 96.4                           & 0.96            & 0.69                           & 0.52                           & 0.8                            \\
                  & AdaGGI($\mathcal{E}_{LUCB}$)                                                                             & 98.4                           & 0.98            & 0.63                           & 0.48                           & 0.52                           & 96.7                           & 0.97            & 0.68                           & \textbf{0.51}                           & 0.57                           \\
                  & AdaGGI($\mathcal{E}_{unif}$)                                                                             & 96                             & 0.96            & 0.64                           & 0.61                           & \textbf{0.15} & 90.2                           & 0.90            & 0.70                           & 0.64                           & \textbf{0.16} \\
                  & AdaGCPI($\mathcal{R}_{fut}$)                                                                             & 96                             & 1.05            & 0.63                           & 0.62                           & \textbf{0.15} & 91.9                           & 0.994           & 0.68                           & 0.66                           & \textbf{0.16} \\
                  & \begin{tabular}[c]{@{}l@{}}AdaGCPI\\ ($\mathcal{R}_{fut} + \mathcal{R}_{pop\textrm{-}fut}$)\end{tabular} & 95                             & 1.04            & \textbf{0.61} & 0.61                           & \textbf{0.15} & 92                             & 0.97            & \textbf{0.67} & 0.65                           & \textbf{0.16} \\ \midrule
C:$[0, 0.1, 0.3]$   & GSDS                                                                                                     & \textbf{100}  & 2.03            & \textbf{0.50} & 0.50                           & 0.50                           & \textbf{100}  & 1.98            & \textbf{0.51} & 0.51                           & 0.5                            \\
                  & AdaGGI($\mathcal{E}_{LCB}$)                                                                              & 99                             & 1.00            & 0.55                           & {0.29} & 0.59                           & 79                           & 0.87            & 0.93                           & {0.34} & 0.57                           \\
                  & AdaGGI($\mathcal{E}_{UCB}$)                                                                              & 
                  \textbf{100}                            & 1.09            & 0.90                           & \textbf{0.25}                           & 0.81                           & \textbf{100}                            & 1.08            & 0.93                           & \textbf{0.29}                           & 0.87                           \\
                  & AdaGGI($\mathcal{E}_{LUCB}$)                                                                             & 99.9                           & 1.08            & 0.90                           & 0.26                           & 0.83                           & \textbf{100}                            & 1.06            & 0.93                           & \textbf{0.29}                           & 0.85                           \\
                  & AdaGGI($\mathcal{E}_{unif}$)                                                                             & 99.6                           & 1.06            & 0.91                           & 0.45                           & 0.53                           & 98.5                           & 1.03            & 0.96                           & 0.65                           & 0.59                           \\
                  & AdaGCPI($\mathcal{R}_{fut}$)                                                                             & 99.3                           & 2.28            & 0.55                           & 0.55                           & 0.53                           & 97.7                           & 2.25            & 0.6                            & 0.59                           & 0.47                           \\
                  & \begin{tabular}[c]{@{}l@{}}AdaGCPI\\ ($\mathcal{R}_{fut} + \mathcal{R}_{pop\textrm{-}fut}$)\end{tabular} & 89                             & 2.28            & 0.55                           & 0.55                           & \textbf{0.44} & 0.98                           & 2.26            & 0.59                           & 0.59                           & \textbf{0.46} \\ \midrule
D:$[0.2, 0.2, 0.2]$ & GSDS                                                                                                     & \textbf{100}  & 2.98            & 0.50                           & 0.5                            &                                & \textbf{100}  & 2.97            & 0.5                            & 0.5                            &                                \\
                  & AdaGGI($\mathcal{E}_{LCB}$)                                                                              & 99.8                           & 2.27            & 0.94                           & \textbf{0.36} &                                & 99.7                           & 2.06            & 0.96                           & \textbf{0.4}  &                                \\
                  & AdaGGI($\mathcal{E}_{UCB}$)                                                                              & 93.8                           & 2.02            & 0.94                           & 0.53                           &                                & 91.4                           & 1.81            & 0.96                           & 0.53                           &                                \\
                  & AdaGGI($\mathcal{E}_{LUCB}$)                                                                             & 95.9                           & 2.07            & 0.94                           & 0.51                           &                                & 93.5                           & 1.83            & 0.96                           & 0.54                           &                                \\
                  & AdaGGI($\mathcal{E}_{unif}$)                                                                             & 83                             & 1.76            & 0.94                           & 0.65                           &                                & 75.1                           & 1.48            & 0.96                           & 0.65                           &                                \\
                  & AdaGCPI($\mathcal{R}_{fut}$)                                                                             & 99.7                           & 2.97            & 0.37                           & 0.37                           &                                & 99.7                           & 2.99            & 0.41                           & 0.41                           &                                \\
                  & \begin{tabular}[c]{@{}l@{}}AdaGCPI\\ ($\mathcal{R}_{fut} + \mathcal{R}_{pop\textrm{-}fut}$)\end{tabular} & 99.8                           & 2.99            & \textbf{0.37} & 0.37                           &                                & 99.7                           & 2.98            & \textbf{0.41} & \textbf{0.4}  &                                \\\midrule
E:$[0.3, 0.3, 0.3]$ & GSDS                                                                                                     & 100                            & 3               & 0.5                            & 0.5                            &                                & 100                            & 3               & 0.5                            & 0.5                            &                                \\
                  & AdaGGI($\mathcal{E}_{LCB}$)                                                                              & 100                            & 3               & 0.49                           & \textbf{0.16} &                                & 100                            & 3               & 0.53                           & \textbf{0.18} &                                \\
                  & AdaGGI($\mathcal{E}_{UCB}$)                                                                              & 100                            & 3               & 0.49                           & 0.25                           &                                & 100                            & 3               & 0.53                           & 0.26                           &                                \\
                  & AdaGGI($\mathcal{E}_{LUCB}$)                                                                             & 100                            & 3               & 0.49                           & 0.24                           &                                & 100                            & 3               & 0.53                           & 0.26                           &                                \\
                  & AdaGGI($\mathcal{E}_{unif}$)                                                                             & 100                            & 3               & 0.49                           & 0.33                           &                                & 100                            & 3               & 0.53                           & 0.34                           &                                \\
                  & AdaGCPI($\mathcal{R}_{fut}$)                                                                             & 100                            & 3               & 0.17                           & 0.17                           &                                & 100                            & 3               & 0.18                           & \textbf{0.18}                           &                                \\
                  & \begin{tabular}[c]{@{}l@{}}AdaGCPI\\ ($\mathcal{R}_{fut} + \mathcal{R}_{pop\textrm{-}fut}$)\end{tabular} & 100                            & 3               & \textbf{0.17} & 0.17                           &                                & 100                            & 3               & \textbf{0.18} & \textbf{0.18} &         \\\bottomrule                      
\end{tabular}
\vspace{1mm}

\end{threeparttable}
\end{center}
\caption{Results for simulated clinical trials with binary outcomes (left) and normal outcomes (right) using different treatment effect vectors $\theta$; averaged across 1000 replications.\\
\small \textit{Column legend:} (1) \%Succ. : prop. of trials which found a significant effect in \textit{some} group. (2) $|\mathcal{S}|$: Average size of discovered subpopulation $\mathcal{S}$. (3) $t_{stop}/B$: Average algorithm termination time (as prop. of budget). (4) $t_{1g}/B$: Average time it took to identify the \textit{first} good arm (as prop. of budget). (5) $t_{1b}/B$: Average time it took to discard the \textit{first} bad arm (as prop. of budget).}\label{tab:trialres-full}
\end{table}

\newpage
\end{document}